\documentclass[a4paper,fleqn]{cas-sc}

\usepackage{lineno}
\usepackage[noheads,nolists]{endfloat}

\usepackage{caption}
\usepackage[labelformat=simple]{subcaption}

\usepackage{url}
\usepackage{multirow}
\usepackage{booktabs}
\usepackage{tabularx}

\usepackage{ragged2e}
\usepackage{makecell}
\usepackage{CJK}
\usepackage{longtable}
\usepackage{csquotes}
\usepackage{titlesec}
\usepackage{placeins}
\usepackage[authoryear]{natbib}

\usepackage{graphicx}
\usepackage{float} 
\def\tsc#1{\csdef{#1}{\textsc{\lowercase{#1}}\xspace}}
\tsc{WGM}
\tsc{QE}
\tsc{EP}
\tsc{PMS}
\tsc{BEC}
\tsc{DE}
\definecolor{lth_color}{RGB}{0,0,0}
\definecolor{ww_color}{RGB}{0,0,0}
\newcommand{\lth}[1]{\textcolor{lth_color}{#1}}
\newcommand{\ww}[1]{\textcolor{ww_color}{#1}}

\begin{document}
\let\WriteBookmarks\relax
\def\floatpagepagefraction{1}
\def\textpagefraction{.001}

\shorttitle{A review on vision-based analysis for automatic dietary assessment}    

\shortauthors{Wei Wang \emph{et al.}}  

\title [mode = title]{A review on vision-based analysis for automatic dietary assessment}



%

\author[1,3,4]{Wei Wang}[
    style=chinese
]
\ead{wwei@st.btbu.edu.cn}

\author[2,5]{Weiqing Min}[
    style=chinese,
    orcid=0000-0001-6668-9208
]
\ead{minweiqing@ict.ac.cn}
\ead[url]{https://vipl.ict.ac.cn/homepage/minweiqing/Home.html}
\cormark[1]

\author[2,5]{Tianhao Li}[
    style=chinese,
    orcid=0000-0001-5220-5743
]
\ead{litianhao21@mails.ucas.ac.cn}

\author[1,3,4]{Xiaoxiao Dong}[
    style=chinese
]
\ead{2030701002@st.btbu.edu.cn}

\author[1,3,4]{Haisheng Li}[
    style=chinese,
    orcid=0000-0003-4861-0513
]
\ead{lihsh@th.btbu.edu.cn}
\ead[url]{https://cse.btbu.edu.cn/szdw/xbxx/rjgcx/176548.htm}
\cormark[1]

\author[2,5]{Shuqiang Jiang}[
    style=chinese
]
\ead{sqjiang@ict.ac.cn}






\affiliation[1]{organization={School of Computer and Engineering, Beijing Technology and Business University},
            city={Beijing},
            postcode={100048}, 
            country={China}}

\affiliation[2]{organization={The Key Laboratory of Intelligent Information Processing, Institute of Computing Technology, Chinese Academy of Sciences},
            city={Beijing},
            postcode={100190}, 
            country={China}
            }

\affiliation[3]{organization={Beijing Key Laboratory of Big Data Technology for Food Safety},
            city={Beijing},
            postcode={100048}, 
            country={China}}
            
\affiliation[4]{organization={National Engineering Laboratory For Agri-product Quality Traceability},
            city={Beijing},
            postcode={100048}, 
            country={China}}

\affiliation[5]{organization={University of Chinese Academy of Sciences},
            city={Beijing},
            postcode={100049}, 
            country={China}}



\cortext[1]{Corresponding author}



\begin{abstract}
\noindent\emph{Background:}
Maintaining a healthy diet is vital to avoid health-related issues, e.g., undernutrition, obesity and many non-communicable diseases. An indispensable part of the health diet is dietary assessment. \ww{Traditional manual recording methods are not only burdensome but time-consuming, and contain substantial biases and errors.}
Recent advances in Artificial Intelligence \ww{(AI)}, especially computer vision technologies, have made it possible to develop automatic dietary assessment solutions, which are more convenient, less time-consuming and even more accurate to monitor daily food intake.

\noindent\emph{Scope and approach:}
\ww{This review presents Vision-Based Dietary Assessment (VBDA) architectures, including multi-stage architecture and end-to-end one. The multi-stage dietary assessment generally consists of three stages: food image analysis, volume estimation and nutrient derivation. The prosperity of deep learning makes VBDA gradually move to an end-to-end implementation, which applies food images to a single network to directly estimate the nutrition. The recently proposed end-to-end methods are also discussed. We further analyze existing dietary assessment datasets, indicating that one large-scale benchmark is urgently needed, and finally highlight critical challenges and future trends for VBDA.}

\noindent\emph{Key findings and conclusions:}
After thorough exploration, we find that multi-task end-to-end deep learning approaches are one important trend of VBDA. Despite considerable research progress, many challenges remain for VBDA due to the meal complexity. We also provide the latest ideas for future development of VBDA, e.g., fine-grained food analysis and accurate volume estimation. \ww{This review aims to encourage researchers to propose more practical solutions for VBDA.}
\end{abstract}



\titleformat*{\section}{\large\bfseries}
\titleformat*{\paragraph}{\normalsize\bfseries}

\maketitle

\setquotestyle{english}
\section{Introduction} 
The malnutrition such as undernutrition, overweight and obesity is now increasingly recognised as one of the greatest health and societal challenges \citep{ingram2020future}. 
\ww{In 2020, 39 million children under 5 years old were overweight or obese due to unhealthy diets, such as the intake of high-fat and high-energy food \citep{WHO}. 
Recently, \cite{cao2021mesenteric} found that a high-fat diet can damage the mesenteric lymphatic vessels, and the leakage of lymphatic vessels can cause metabolic problems such as abdominal obesity and insulin resistance.
It is worth mentioning that the malnutrition is a major risk factor for many non-communicable diseases, such as cardiovascular disease, diabetes, and certain cancers \citep{lauby2016body}.
The previous International Agency for Research on Cancer (IARC) working group concluded that there was sufficient evidence for a cancer-preventive effect of avoidance of weight gain for cancers of the colon, kidney (renal cell), breast (postmenopausal), esophagus (adenocarcinoma), and corpus uteri \citep{international2002iarc}.}
Fortunately, obesity and many chronic diseases can be prevented via dietary assessment, which can monitor daily food intake and control eating habits~\citep{nordstrom2013food}. Furthermore, we can utilize these dietary data to analyze the relations between dietary patterns and some diseases~\citep{chu2021dietary}, and support personalized nutrition~\citep{mcdonald2016personalized}. Therefore, dietary assessment  has become the focus of widespread attention in various fields of computer vision, medicine, nutrient and health ~\citep{DBLP:journals/corr/abs-2103-03375,mcpherson2000dietary,DBLP:journals/tsc/LiuCLCVMCH18,probst2015dietary,DBLP:journals/tim/PouladzadehSA14}.

Over the years, researchers have explored various methods for dietary assessment, such as the \ww{24-h-dietary recall (24-HDR) \citep{slimani2000standardization, foster2008children, kirkpatrick2014performance}, food frequency questionnaire (FFQ) \citep{willett1985reproducibility,  wong2008evaluation, forster2014online,  kristal2014evaluation},  dietary record (DR) \citep{gersovitz1978validity}}, and brief dietary assessment instruments or screeners \citep{illner2012review}. Among them, FFQ can be regarded as a long-term dietary assessment approach, while 24-HDR and dietary records are the primary subjective methods used in short-term evaluations \citep{shim2014dietary}. The implementation of these methods mainly involves paper questionnaires and interview-based tools \citep{thompson2015comparison}, which have provided an important contribution to the development of nutrition research. Nevertheless, these traditional manual recording methods are time-consuming, inaccurate~(e.g., energy intake under-reporting \citep{gibney2020uncertainty}), labor-intensive, and high-demanding for a certain level of literacy and communication skills. 
Therefore, traditional methods make the research process more difficult or even infeasible for special populations such as children, adolescents and the elderly. In addition, these issues make the collected dietary data unreliable and difficult to conduct large-scale assessment. To overcome these problems, some automatic methods have been developed, such as smart cooking systems and portable systems~\citep{DBLP:journals/imm/PouladzadehSY16}. The major limitations to these approaches are  the limited or illogical usage. For example, the smart kitchen~\citep{chi2008enabling} captures images of the food preparation process to measure all of the ingredients inside the kitchen, yet has  the inability to be used outside the home. 
\cite{gobe} only calculates the calories after the user has already eaten the food via measuring the glucose of  user’s cells, which is late for patients who need to know the amount of calories before eating.

Recent advances in Artificial Intelligence (AI), especially computer vision and machine learning, have paved the way for more robust automatic dietary assessment. The widespread use of portable devices (e.g., smartphones) with enhanced capabilities together with the advancements in computer vision enabled the development of VBDA. \ww{It allows users to easily measure the nutrition and calories of food by taking pictures of food through their mobile or wearable devices. Compared with traditional methods, VBDA can provide a solution to eliminate subjectivity, get rid of time and space constraints, and enhance the comprehensiveness and accuracy of dietary intake assessment. It can not only reduce the burden of keeping food journaling, but also provide immediate dietary assessments, demonstrating great potentials in effective diet monitoring and control. Furthermore, driven by advances in deep learning \citep{lecun2015deep}, the higher capacity of automatic food visual analysis results in higher performance of dietary assessment. For these reasons, VBDA is becoming one mainstream method for dietary assessment. It should be emphasized that deep learning algorithms become computationally expensive when applied to high-dimensional data (e.g., food images), possibly due to the learning phase associated with a deep-layered hierarchy \citep{DBLP:journals/imm/PouladzadehSY16}.}

\ww{Some surveys related to VBDA have been conducted. For example, \cite{tay2020current} focused on highlighting various methods of food volume estimation. We pointed out that visual analysis and volume estimation are the basis of multi-stage dietary assessment. \cite{liu2021efficient} reviewed Convolutional Neural Networks (CNNs) for feature extraction from food images. We summarized both deep learning methods and traditional manual feature extraction methods (e.g., SIFT and LBP) for food visual analysis. Additionally, other articles focused on surveying various food recognition methods \citep{DBLP:journals/access/SubhiAM19, DBLP:journals/titb/LoSQL20, knez2020food, DBLP:journals/corr/abs-2106-11776}, where most of these recognition works are not evaluated on the dietary assessment framework but only on the recognition task. \cite{hochsmann2020review} illustrated only 12 different image-assisted food intake assessment methods. In contrast, we have listed more methods to analyze their effectiveness and feasibility comprehensively. 
In summary, different from their works, we proposed two VBDA architectures to more comprehensively and systematically review current dietary assessment methods, and discussed their advantages and limitations. In addition, most of our surveyed works are directly for VBDA. Finally, we discuss existing VBDA datasets and evaluation metrics, and the key challenges and future trends for VBDA are highlighted.}

\ww{This review is organized into the following subtopics:} Section \ref{VBDA} introduces the VBDA framework including multi-stage one and end-to-end one. Section \ref{MSDA} describes the multi-stage dietary assessment including vision-based food analysis, food portion estimation and \ww{nutrient derivation}. The end-to-end approaches are summarized in Section \ref{E2EA}. We present various dietary assessment datasets and evaluation metrics in Section \ref{Datasets}. Section \ref{FutureWorks} discusses future development trends in this field. Finally, we conclude this article in Section \ref{Conclusion}.

\section{Vision-based dietary assessment}
\label{VBDA}
\begin{figure}[htbp]
    \centering
    \begin{subfigure}{\textwidth}
        \centering
        \includegraphics[width=0.9\textwidth]{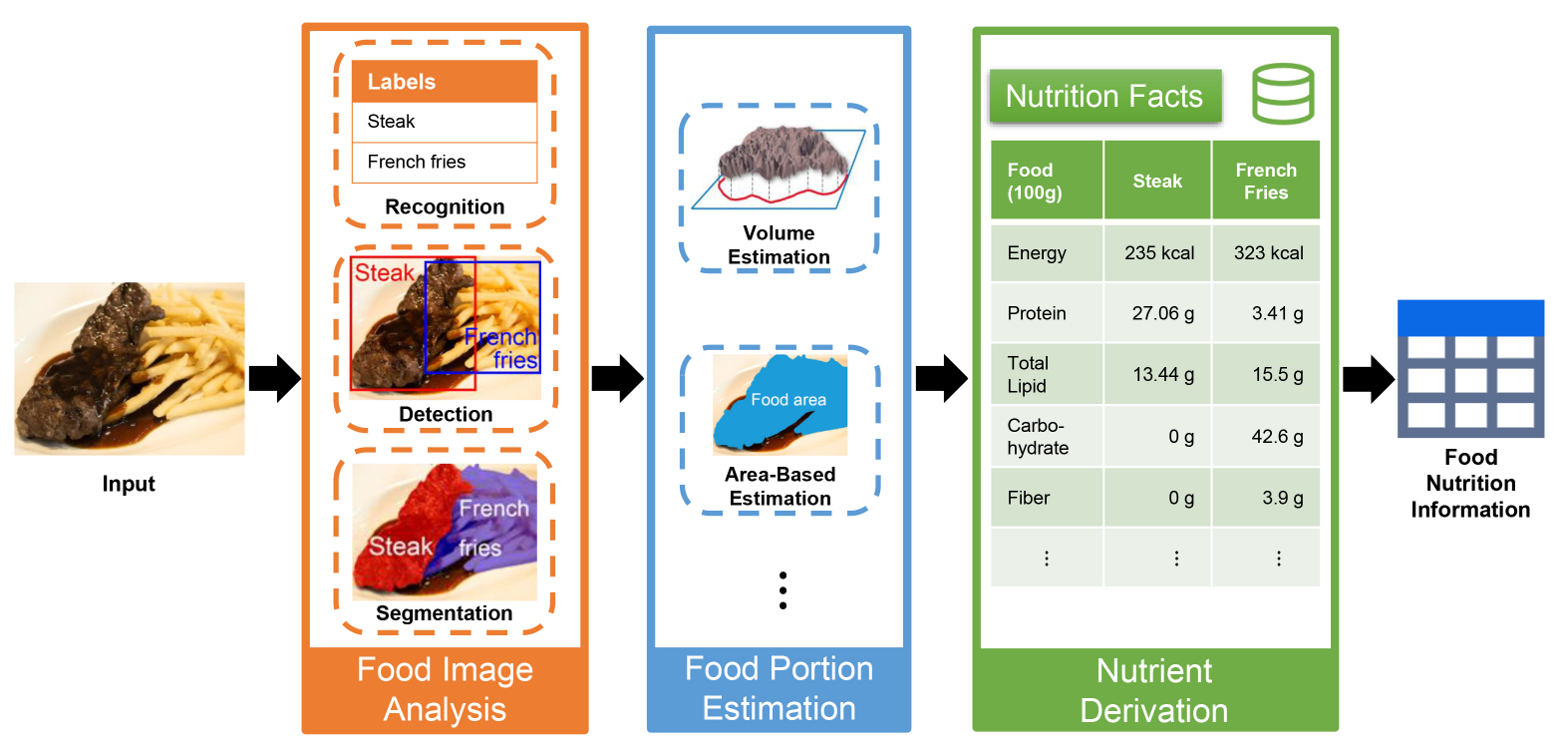}
        \caption{Multi-stage architecture for dietary assessment.}
        \label{multistage}
    \end{subfigure}
    \\
    \begin{subfigure}{\textwidth}
        \centering
        \includegraphics[width=0.95\textwidth]{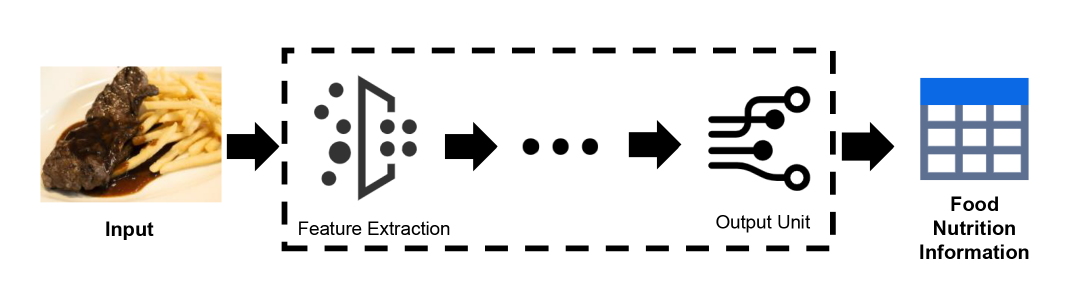}
        \caption{End-to-end architecture for dietary assessment.}
        \label{endtoend}
    \end{subfigure}
    \caption{\ww{The two architectures of VBDA: \subref{multistage} The multi-stage dietary assessment, which mainly consists of food image analysis, portion estimation, and nutrient derivation. \subref{endtoend} The end-to-end architecture emphasizes a single model instead of a pipeline of multi-stage methods. Only the original inputs and the final outputs are required to be specified, while the information learned by the neural network is internally relevant.}}  
    \label{vbda-archs}
\end{figure}

\ww{VBDA refers to taking images of a meal as input and then using computer vision to automatically identify relevant dietary information as the output. We divide it into two architectures, as shown in Figure \ref{vbda-archs}. In the earlier years, the VBDA was conducted in a step-by-step way, namely multi-stage dietary assessment. In recent years, with the rapid development of deep learning, instead of training a pipeline of models to handle subtasks at different stages, the end-to-end deep learning solutions for dietary assessment are proposed to apply input data to a single network for direct nutrient derivation. Next, we will briefly introduce multi-stage and end-to-end solutions, respectively. At the end of this section, we summarize the representative works of the VBDA.}



\subsection{Multi-stage architecture}
\ww{As shown in Figure \ref{multistage}, the multi-stage VBDA generally consists of three stages, including food image analysis, portion estimation, and nutrient derivation. The performance of the first two stages highly depends on the used AI algorithms and the available food datasets, while the last stage depends on the nutritional composition database. }

\paragraph{Food image analysis.}
The key in the visual analysis is compact and expressive feature representations that summarize  the information of food images. This phase is related to food recognition, detection and segmentation.  The first column of Figure \ref{multistage} shows one example for food recognition, detection and segmentation. Food recognition mainly predicts the type or composition of food items from images \citep{DBLP:journals/csur/MinJLRJ19}. Food detection aims to locate and classify each food item from the food image, where the localization is realized via estimating the bounding box of each dish. Compared with food recognition, food detection additionally provided the localization for the recognized food item. Food segmentation involves partitioning food images into multiple food items in a pixel-level way.  It is the process of assigning a food label to every pixel in one food image. Therefore, it is more precise to localize food items with arbitrary shapes compared with rectangular shapes for food detection. Such more precise food area localization is generally more helpful for the following  portion estimation of these food items.

\paragraph{Food portion estimation.}
The visual features are far from sufficient to estimate calorie content. In order to accurately estimate the nutrient content of the food from the image, it is necessary to identify the food item and estimate the weight. Earlier methods conduct rough food weight estimation based on food area \citep{DBLP:conf/icmcs/HeXKBD13, DBLP:conf/mm/OkamotoY16}. As the technology evolves, calculating the food portion based on volume estimation in combination with food density databases becomes more popular.

\paragraph{Nutrient derivation.}
The conversion from food to meaningful nutritional information depends on the accuracy of the nutrition fact databases.\lth{There are several food databases in the public domain, such as USDA Food and Nutrient Database for Dietary Studies (FNDDS) \citep{FNDDS}, Food Databanks National Capability extended dataset (FDNC) \citep{FDNC} , The Canadian Nutrient File \citep{CNF} and Open Food Facts \citep{OFF}.}
\ww{Take FNDDS as an example, as shown in Table \ref{FNDDStable}. The FNDDS contains the most common foods and beverages consumed in the United States, their nutritional values, and the weights of typical food portions. Nutritional value data for foods are stored in these tables, which are available from national and international health organizations. The information in the nutrition database for each food can be used to calculate the energy of the food consumed. The actual nutrient content $N$ can be calculated as:}

\begin{equation}
    N=\;\frac{N_T\times M}{M_T} 
\end{equation}
where the food mass $M$ can be obtained after portion estimation. $N_T$ and $M_T$ represent food nutrition and mass in the table, respectively. \ww{Therefore, the nutrient content of each food consumed is obtained by simply subtracting the nutrient content after the meal from the nutrient content before the meal.} This means that the energy and nutrients of the food a person eats can be assessed based on the images obtained before and after the food is consumed \citep{DBLP:conf/cimaging/WooOKEDB10}.

\begin{table}[htbp]
        \centering

        \caption{A sample of USDA Food and Nutrient Database for Dietary Studies (FNDDS).}\label{FNDDStable}
        \begin{tabular*}{\tblwidth}{@{}@{\extracolsep{\fill}}lccccc@{}}
         \toprule
         \centering Food (100g)  & Energy (kcal) & Protein (g) & Carbohydrate (g) & Fiber (g) & Sugars (g)\\
         \midrule
            Apple, raw                     & 52  & 0.26  & 13.81 & 2.4 & 10.39\\
            Orange, raw                    & 47  & 0.94  & 11.75 & 2.4 &  9.35\\
            Tomatoes, raw                  & 18  & 0.88  &  3.89 & 1.2 &  2.63\\
            Bread, white                   & 270 & 9.43  &  49.2 & 2.3 &  5.34\\
            Egg, whole, raw                & 143 & 12.56 &  0.72 &   0 &  0.37\\
            Cucumber, raw                  & 15  & 0.65  &  3.63 & 0.5 &  1.67\\
            Banana, raw                    & 89  & 1.09  & 22.84 & 2.6 & 12.23\\
            Orange, raw                    & 47  & 0.94  & 11.75 & 2.4 &  9.35\\
         \bottomrule
        \end{tabular*}
       
\end{table}

\subsection{End-to-end architecture}



\ww{Although multi-stage VBDA improves a number of baseline methods for dietary assessment, they have some limitations. First, these methods are required to be defined and optimized individually at each stage, while their accuracy remains a challenge \citep{DBLP:journals/corr/abs-2011-01082}. Since the error propagation path leads to the accumulation of errors, multi-stage leads to the accumulation of errors and has an impact on subsequent operations. The end-to-end architecture replaces these different stages with a single neural network, reducing error propagation and joint optimization paths. Second, the multi-stage architecture requires the definition of separate stages as well as inputs and outputs for each stage. It means that potentially helpful information from the early stages cannot be passed on or used later to improve predictions. In contrast, the end-to-end architecture emphasizes a single model instead of a pipeline of multi-stage methods. Only the original inputs and the final outputs are required to be specified, while the information learned by the neural network is internally relevant. Third, the approach mainly relies on pixel-wise annotations of large amounts of data and additional information in the food images, such as volume. Based on a comprehensive discussion of existing works, we articulate key open challenges and provide some prospective analysis of possible VBDA solutions in Section \ref{FutureWorks}. Considering the nutrient estimation involves multiple types of nutrient components, such as proteins and fat, a multi-task end-to-end framework for nutrient estimation is generally adopted. Figure \ref{endtoend} shows its overall framework.}


\subsection{The timeline of VBDA}
\begin{figure}[htbp]               
        \centering
        \includegraphics[width=\textwidth]{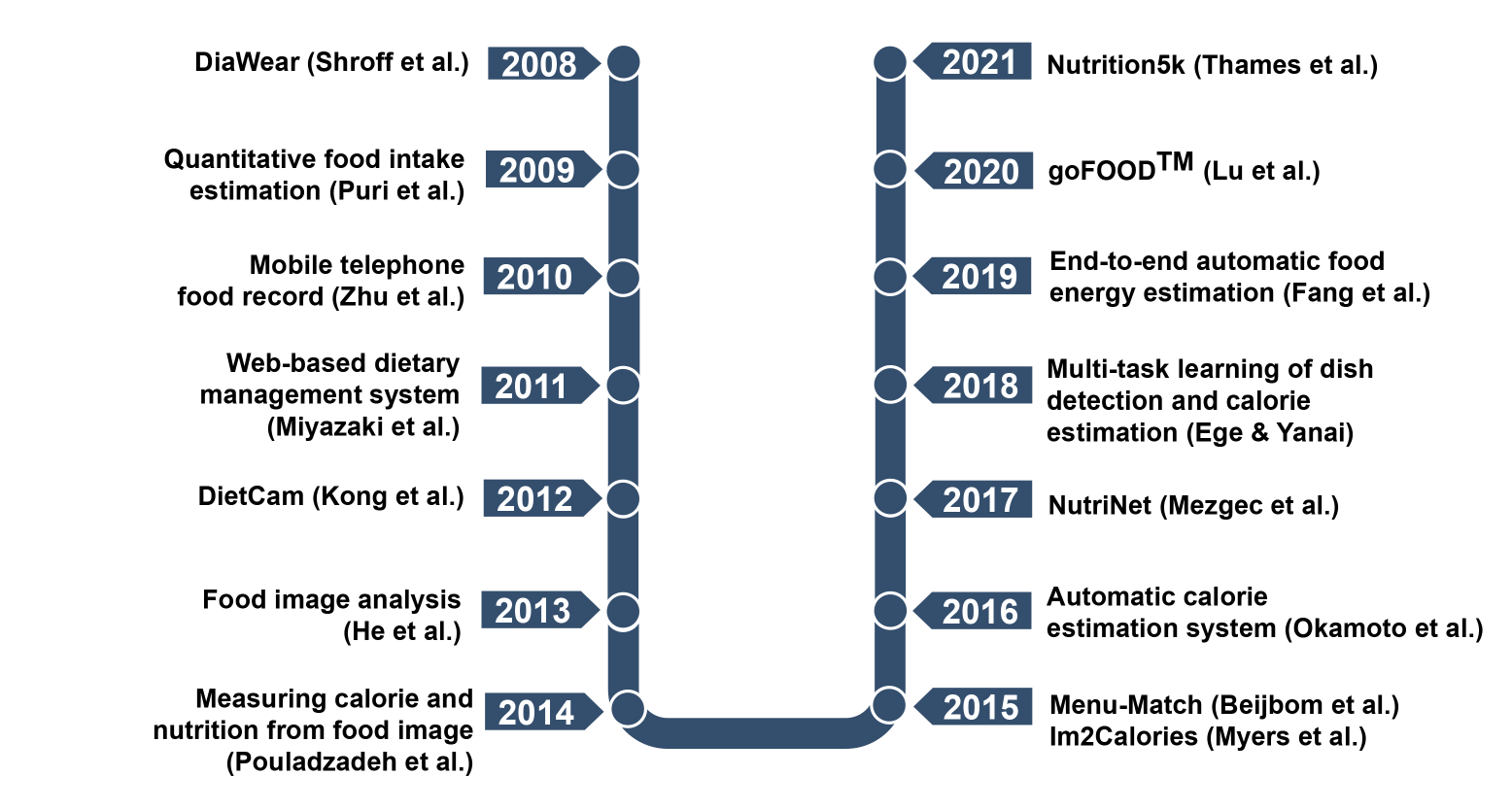}
        \caption{Representative works for VBDA.}
        \label{timeline}
\end{figure}

Throughout the history of VBDA development, we have summarized some representative studies as shown in Figure \ref{timeline}.
To the best of our knowledge, \cite{DBLP:conf/iswc/ShroffSS08} first conducted a vision-based method for food recognition, and the calorie of recognized food items is then obtained from a lookup table. Similar calorie lookup methods have also been adopted by \cite{DBLP:journals/tim/PouladzadehSA14}. Considering the food portion was not estimated, the above-mentioned methods can not obtain the estimated calorie of the whole food.  In comparison, \cite{DBLP:conf/wacv/PuriZYDS09} used 3D reconstruction for volume estimation after identifying each food item, realizing the first quantitative diet assessment application. \cite{zhu2010image} developed a mobile food recording system by performing food visual analysis, volume estimation, and nutrient estimation continuously, giving birth to the rudiments of subsequent multi-stage VBDA.
\cite{miyazaki2011image} explored a new approach to estimate calorie without food recognition and volume estimation. Multiple food features are extracted and then searched in a dictionary dataset, inspiring the development of end-to-end approaches. DietCam \citep{DBLP:journals/percom/KongT12}  processed the food image through a multi-view understanding  and estimated the volume ratio through the size of credit card, which is placed next to the plate.  \cite{DBLP:conf/icmcs/HeXKBD13} further took into account the shape of the food, and   developed an area-based method to estimate the weight of irregularly-shaped foods. Two works  Menu-Match~\citep{DBLP:conf/wacv/BeijbomJMSK15} and Im2Calories~\citep{DBLP:conf/iccv/MyersJRKGSGPH015} focused on the restaurant scenario for calorie estimation from a single image. The difference is that Menu-Match combined different hand-crafted features while Im2Calories adopted CNNs for feature extraction for its powerful representation. Later, more advanced CNNs have been developed for dietary assessment~\citep{DBLP:conf/mm/OkamotoY16,mezgec2017nutrinet}. \cite{DBLP:conf/ijcai/EgeY18}  achieved nutrient assessment of multiple dishes using multi-target detection methods. \cite{fang2019end} performed the end-to-end \ww{nutrient estimation}  by generating energy distribution maps of food images through the generative adversarial network. \cite{goFOODTM} proposed a nutrient assessment system using two images. Its accurate estimation performance and friendly user interface showed the possibility of applying VBDA to daily life. To produce higher performance, recently, \cite{DBLP:journals/corr/abs-2103-03375} utilized multi-sensor RGB-D food images for end-to-end VBDA, and meanwhile released the nutritional dataset Nutrition5k. 

\section{Multi-stage dietary assessment}
\label{MSDA}
\subsection{Food image analysis}
\label{FIA}
In the multi-stage work, food image analysis aims to provide necessary information for the later stages, among which the essential information is the category of food and its visual regions. 
Many works at this stage established different food image analysis pipelines. According to the output types, we classify them into food recognition, food detection, and food segmentation. 
\ww{Given that various methods for food image analysis were described, we have added Table \ref{FIAtable} to compare the performance of different approaches. The table lists different methods, dataset sizes, method evaluation, and nutrient estimation. Since most methods use their own datasets, we choose to list the size of the datasets to help readers have a general understanding. For common nutrition-related datasets and evaluation metrics mentioned in the table, we describe them in detail in Section \ref{Datasets}.}


\renewcommand{\~}{$\sim$}

\newcolumntype{P}[1]{>{\Centering\hspace{0pt}}p{#1}}
\newcolumntype{F}[1]{>{\hspace{0pt}}p{#1}}
\newcolumntype{Z}{>{\centering\let\newline\\\arraybackslash\hspace{0pt}}X}
\begin{table}[htbp]
\renewcommand\arraystretch{1.3}
    \caption{\lth{Food image analysis approaches in multi-stage VBDA.}} \label{FIAtable}
\scriptsize
     \begin{tabularx}{\textwidth}{F{3.8cm}F{2.2cm}F{4.5cm}F{1.5cm}F{2.6cm}}
    \toprule
    \textbf{References} &  \textbf{Dataset\# images(class)} &  \textbf{Methods} & \textbf{Method \quad Evaluation} & \textbf{Nutrient Estimation}\\
    \midrule
    
    \textbf{Recognition} & & & & \\
    \cite{DBLP:conf/icmcs/WuY09} & 100(\~10)  & SIFT feature & 52\% & - \\
    \cite{DBLP:journals/percom/KongT12} &-(-)& SIFT feature and nearest neighbor classifier & \~92\% &  \\
    \cite{DBLP:conf/wacv/PuriZYDS09} & -(\~150) & Color neighborhood, maximum response features and SVM classifier & \~90\% & 5.75($\pm$ 3.75)\% on volume estimation\\
    \cite{DBLP:conf/siggrapha/ChenYHWLCYO12} & 100(50) & Multi-feature fusion, SVM classifier & 68.3\%& - \\
    \cite{DBLP:conf/aaai/ChokrE17} & \~1,000(6)&  Random forest and SVM classifier& 99.1\% &MAE 0.0933 on calorie estimation\\
    %
    \cite{christodoulidis2015food} & 246(573) & Own designed CNN & 84.9\%& - \\
    \cite{mezgec2017nutrinet} & 225,953(520) & Modification of the AlexNet & 86.72\% & -  \\
    \cite{ayon2021foodiecal} &23,000(23)&InceptionV3 &89.48\%&-\\
    \cite{ma2021image} & 10,047(100)& InceptionV3 & 78.26\%  & $R^{2}$ 0.73 on protein estimation \\
    \midrule
    \textbf{Detection} & & & & \\
    \cite{DBLP:conf/iciap/CioccaNS15}& 2,000(15)& Circle Hough transform& - & - \\
    \cite{DBLP:journals/tomccap/PouladzadehS17}&7,000(30)&Selective search& 93.05\% Precision& - \\
    \cite{jiang2020deepfood}&12,740(100)&Faster R-CNN&17.5\% mAP&-\\
    \cite{mao2021visual}&14,991(82)&Faster R-CNN&0.7064 F1&-\\
    \citep{DBLP:conf/mmm/OkamotoY16}&-&Background subtraction and the hough transform&-&-\\
    \cite{DBLP:conf/bsn/QiuLL19}&4,200(14)&Mask R-CNN&-&-\\
    \citep{lei2021assessing}&14(-)&Mask R-CNN&-&- \\
    
    \midrule
    \textbf{Segmentation} & & & & \\
    \cite{DBLP:conf/cimaging/ZhuMBKLED08} & -(-) & Threshold segmentation & - & -\\
    \cite{DBLP:conf/cimaging/MariappanBZBKED09} & 50(\~3) & Threshold segmentation& -  & - \\
    \cite{DBLP:conf/aaaifs/EskinM12} & 676(49)& Threshold segmentation with the morphological method& - & - \\
    \cite{zhu2010image} & 63(3) & Normalized cut segmentation & - & Percentage of misreported nutrient information: 1\% \\
    \cite{DBLP:conf/icmcs/HeXKBD13} & 1,453(96) & Local variation segmentation & - & \~10\% on weight estimation\\                  
    \cite{DBLP:journals/titb/ZhuBKBD15} & -(63)&Multiple hypothesis segmentation & - & - \\
    \cite{DBLP:conf/memea/PouladzadehSY14} &-(15) &Graph cut segmentation & \~90\% PA&- \\
    \cite{anthimopoulos2015computer} & 1620(248) & Region growing segmentation& 88.5\% PA   & 10\% MAE on CHO estimation \\
    \cite{DBLP:conf/huc/SudoSMT14} & 2,500(29) & Super-pixel segmentation& - &31.8\% average error on calorie estimation \\
    \cite{DBLP:conf/iccv/MyersJRKGSGPH015} &12,625(201) & Finetuned DeepLab & 25\% IoU & - \\
    \cite{DBLP:conf/i2mtc/PouladzadehKPYS16} &-(30) &Graph cut segmentation & -& -\\
    \cite{DBLP:conf/mm/OkamotoY16} & \~1M(1K) & GrabCut segmentation & -  & 0.213 relative average error on calorie estimation\\
    \cite{DBLP:conf/mm/EgeSY19} & 1,000(-) &  U-Net & 84.1\% mIoU & -\\
    \cite{DBLP:conf/mm/AndoECY19} & 5,301(-)& U-Net & 80\% mIoU& - \\
    \cite{DBLP:conf/mm/DehaisAM16}& 821(-) &  CNN based border map generation, seeded region growing segmentation & 88\% PA&- \\
    \cite{DBLP:conf/iciap/DehaisAM15} & -(-) & Seeded region growing segmentation & 80\% $F_{\text{min}}$ & - \\
    \cite{goFOODTM} &21,807(80)& Mask-RCNN & 83.9\% $F_{\text{min}}$ &Pearson correlations 0.69 on protein estimation \\
    \cite{DBLP:journals/corr/abs-2003-08273} & 1,281(521) &  Pretrained dilated ResNet50 & 83.5\% PA &$R^2$ 0.923 on calorie estimation \\
    
    \bottomrule
    \end{tabularx}
\end{table}

\subsubsection{Food recognition}  
Food recognition can be regarded as the most basic work in visual analysis, and food item labels can be obtained via food recognition. Existing research methods are mainly divided into manual feature-based recognition methods and deep learning-based recognition ones for dietary assessment. The former starts with relevant features  manually extracted from images, and the features are then used to create a classification model to recognize the food class. In contrast, the latter  performs end-to-end learning, where a network is given and a classification task will be performed, and it automatically extracts relevant features from food images and learns how to do this automatically.

\paragraph{Manual feature-based recognition methods.}
Manual feature-based recognition methods include feature extraction and classification. Image feature extraction is the key to food image recognition. Manual features vary from simple features, such as color, texture, and shape, to other complex features. Compared with features such as color and texture, the Scale Invariant Feature Transform (SIFT)
descriptor has the advantages of scale, orientation and affine distortion invariance, which can effectively deal with the deformation characteristics of food images. As a result, it is widely used in food image recognition tasks. \cite{DBLP:conf/icmcs/WuY09} formulated food recognition from videos as an image retrieval task using SIFT. Since this study focused on fast food, the calorie information of most fast food is standardized and available to the public. Once the food type is recognized, corresponding calorie information can be directly estimated from a database without considering food portions. The DietCam system \citep{DBLP:journals/percom/KongT12} was an automatic multi-view food classifier in a health perception system. It calculated SIFT descriptors for food items. Then three meal images from different views or a short video were used to realize 3D reconstruction. With the knowledge of food types and food scales, calories can be estimated via simple calculation. Considering the complementarity between different manual features, more methods integrate different manual features to improve the recognition performance. \cite{DBLP:conf/wacv/PuriZYDS09} adopted Adaboost-based feature selection to combine color neighborhood features and maximum response features to achieve acceptable recognition rates on a large number of food types. \cite{DBLP:conf/siggrapha/ChenYHWLCYO12} combined SIFT, local binary patterns, color, Gabor filter descriptors for food recognition. They separately trained a Support Vector Machine (SVM) classifier for each type of features and fused them using the AdaBoost algorithm for classification. \cite{DBLP:conf/aaai/ChokrE17}  used the Mathworks Image Processing Toolbox to extract raw features from food images. Then they applied InfoGain, selected the top 6,000 features followed by principal component analysis, and ended up with 23 features per image as their final representation for food recognition and food mass estimation. It is worth mentioning that the method was limited to the surface and background of the food images and can only predict the calorie content of a single food. Hand-engineered features are time-consuming, brittle, and not scalable in practice, especially when it comes to unstructured data such as food images and videos. In contrast, deep learning has become the mainstream method for automatically discovering the discriminative feature representation from raw data. \ww{Recent studies have demonstrated that visual features learned by deep learning are more expressive and robust than handcrafted ones \citep{kamilaris2018deep, cai2020obesity}.}  For this reason, recently proposed VBDA methods adopt deep-learning networks for food vision analysis.

\paragraph{Deep learning based recognition methods.}
\ww{Deep learning, as an exclusively superior framework of contemporary machine learning, was extensively explored and brought a new stage for food image recognition \citep{zhou2019application}.}
The key element of deep learning lies in CNN and its derivative algorithms. \cite{christodoulidis2015food} used a six-layer CNN to classify food images. Image chunking was used to extract a set of overlapping square blocks. The voting determined the type of food in the area. In contrast, \cite{mezgec2017nutrinet} proposed NutriNet for food and beverage recognition based on AlexNet. The input image was expanded from 256 $\times$ 256 to 512 $\times$ 512 pixels, and adding a convolutional layer to obtain hidden features in higher resolution images. The model has been used in practice as one part of a mobile application for dietary assessment of Parkinson's disease patients. However, its limitation was that for images with multiple food items, not every item could be accurately identified. 
\lth{\cite{ayon2021foodiecal} proposed a food detection system based on multi-label recognition by setting a threshold. They also deployed trained models into the webpage to improve the user experience.}
\cite{ma2021image} trained their model by fine-tuning four deep architectures that were pre-trained on the ImageNet dataset. They modified the last fully connected layer of the architecture to output 100 classes. Because the food category has been defined and has a fixed nutritional configuration (e.g., calories, carbohydrates and proteins), the nutrient estimation mainly relies on the classification results. \lth{Besides, they carried out almost the same experiments on their another dataset ChinaMartFood-109 \citep{ma2022application}. Unlike studies that summarized food recognition methods, which only recognized food items in their work, we focused on summarizing approaches that directly employed food recognition for the nutrition assessment task. For more works only for food image recognition, please refer to some surveys, such as \cite{DBLP:journals/csur/MinJLRJ19}.}


\subsubsection{Food detection}    
In dietary assessment, multi-food localization and recognition are the primary detection technique application. \cite{DBLP:conf/iciap/CioccaNS15} proposed an automatic dietary monitoring system for canteen customers based on multi-food detection.
They added a plate localization method to reduce the area to be recognized. Several descriptors are evaluated, such as color-based, statistical, spatial-frequency or spectral, structural, and hybrid. Due to the weekly fixed menus and the control conditions of food image acquisition in the canteen, the canteen scenario is relatively simple. The cafeteria staff provides a fixed amount of food according to the nutrition table, and the system can evaluate the leftovers in the canteen to calculate the amount of food consumed. 
\cite{DBLP:journals/tomccap/PouladzadehS17} further improved their system and used the selective search to generate detection areas for multiple foods.
\ww{However, the selective search that depends on the region is time-consuming and computationally burdensome. To address this problem, many studies have applied Faster R-CNN \citep{7485869} to make region proposals more effective. \cite{jiang2020deepfood} applied the Faster R-CNN model to extract the regions of interests (ROIs) and improve the efficiency of the detection model. Then, the feature maps for food recognition were extracted based on CNN. For each item detected in the food image, they assume that the basic weight of each food item was 400 grams, and then summarized the dietary assessment report for the user. In addition, \cite{mao2021visual} trained the food localization step based on Faster R-CNN by selecting regions containing food items in an input image. The selected food regions were resized to a fixed size and fed into the food classification system. Based on the features extracted from the CNN model, visual clustering of similar food categories was performed to establish a visual-aware hierarchy structure. The final output included bounding boxes and food labels for each food.}

\ww{Different from these works, more studies have been proposed to quantify dietary intake by monitoring users' eating behavior. GrillCam \citep{DBLP:conf/mmm/OkamotoY16} roughly estimated the calorie intake by detecting the moment of delivery, identifying food categories, and detecting candidate food areas. It is suitable for situations where the user has not decided how much food to eat. 
However, not much research has been conducted to address the issue of nutrient assessment in shared food scenarios. Most VBDA methods are either targeted for a single user or limited to laboratory environments.With the booming development of deep learning, Mask R-CNN \citep{he2017mask}, which was improved based on Faster R-CNN, was proposed and applied to human pose recognition, with good results in instance segmentation, object detection, and human keypoints detection tasks. Inspired by this, \cite{DBLP:conf/bsn/QiuLL19} estimated dietary intake based on videos of the eating scenes captured by a 360 camera. In their recorded videos, 2-3 users were sharing food together. They integrated food detection with face recognition and hand tracking technology, and used the fine-tuned Mask R-CNN  to detect 13 food categories and people. Similarly, another approach \citep{lei2021assessing} captured videos of a typical household setting where food is shared, integrating dish detection and body pose to quantify dietary intake. The method was designed with a 4-layer feed-forward neural network to infer the eating status of the subjects, utilizing the dish types and the detected food containers bounding boxes for subsequent food intake assessment for each participant. }

\subsubsection{Food segmentation} 
Food segmentation divides one food image into multiple parts. The segmentation process assigns a label to each pixel so that pixels with the same label have specific characteristics. Once the segmentation map is obtained, portion estimation can be done using the segmented food area or splitting the reconstructed 3D model into different foods. Therefore, segmentation methods are more widely used than food recognition and detection in VBDA. \lth{It should be noticed that many works take segmentation as a part of the whole approach without evaluating the accuracy.}

Early food segmentation works were often based on some assumptions. For example, plates should have certain color or shape, and food cannot be overlapped or blocked. These results in larger differences between the background and the food.  \cite{DBLP:conf/cimaging/ZhuMBKLED08} and \cite{DBLP:conf/cimaging/MariappanBZBKED09} used the threshold segmentation method to process the image. The image is converted into a grayscale image and then thresholded to form a binary image. The food item is segmented by a search algorithm. Additionally, the image can also be converted to the YCbCr color space to identify potential light-colored food items. \cite{DBLP:conf/aaaifs/EskinM12} assumed that plates had uniform white color and elliptical shape, and foods were more colorful than the plate. They performed the threshold segmentation with the morphological method to segment the plate and food in succession.
Different from these works, \cite{zhu2010image} applied the active contour model to segment food images. The model can deform an initial curve to the boundary of the object under some constraints from the image. They discussed the limitation of active contours and chose the normalized cut as their segmentation method because it can measure both the total dissimilarity between different groups and the total similarity within the groups.
Normalized cut is a kind of graph-theoretic segmentation method that treats one pixel from the image as one node of a graph and considers segmentation a graph partitioning problem. There are many other graph-theoretic based segmentation methods. For example, 
\cite{DBLP:conf/icmcs/HeXKBD13} used the image-based local variation method to segment food according to the degree of variability in adjacent regions. \cite{DBLP:journals/titb/ZhuBKBD15} proposed multiple segmentation hypotheses and adopted normalized cut to dynamically select the number of segments guided by the classification results. The similarity between these two works lies in the iterative use of the classification feedback to refine the  segmentation results. A series of works by Pouladzadeh \emph{et al.} used the graph cut in the task of food calorie estimation. \cite{DBLP:conf/memea/PouladzadehSY14} introduced a new segmentation approach that combined the graph cut with the texture segmentation method. It was widely used in their later works \citep{DBLP:journals/tim/PouladzadehSA14}. These works combined the segmentation approach with SVM to analyze food images. GoCARB \citep{anthimopoulos2015computer} is a mobile daily carbohydrate assessment system for patients with type 1 diabetes. The food was segmented using a region-growing algorithm after detecting the localized plate regions. It should be noted that the system has certain assumptions, including the use of flat-bottomed elliptical dishes, and food cannot be overlapped or blocked. It can also be extended to display other macronutrients, advancing the development of a healthy human diet. However, all of the above are unsupervised segmentation methods. In contrast, in another study \citep{DBLP:conf/huc/SudoSMT14}, semantic segmentation completes image segmentation and labeling. Label histogram shows the frequency of occurrence of food ingredient labels, based on which regression analysis was performed, and nutritional data was estimated. However, in order to train the semantic segmentation model, a large amount of recipe data and corresponding food images and ingredient labels need to be collected.

Afterward, the methods with deep learning have become more and more popular. Im2Calories \citep{DBLP:conf/iccv/MyersJRKGSGPH015} was the first work to achieve semantic segmentation specifically for food images using deep CNN. They combined the segmentation network with a deep prediction network and converted voxels to estimate the food volume. However, their evaluation was limited to plastic food replicas for training nutritionists. \cite{DBLP:conf/i2mtc/PouladzadehKPYS16} took full advantage of the graph cut segmentation and CNNs for calorie measurement. However, they could only be applied to images with a single food label. \cite{DBLP:conf/mm/OkamotoY16} applied GrabCut with bounding boxes as a segmentation seed, and used a CNN-based recognition engine to classify them. In addition, food segmentation methods based on region growing and merging algorithms were subsequently improved via boundary detection networks. \ww{DepthCalorieCam \citep{DBLP:conf/mm/AndoECY19} employed food segmentation based on U-Net \citep{ronneberger2015u}, the accuracy of segmentation was much higher than \cite{DBLP:conf/mm/OkamotoY16}.} 
\lth{Similarly, \cite{DBLP:conf/mm/EgeSY19} also used U-Net for segmentation. The difference is that the it estimated food calories based on the actual size of the food area calculated, while the DepthCalorieCam used a depth camera to obtain the food volume information and applied the estimated food volume to a regression equation of the calorie. 
}
\cite{DBLP:conf/mm/DehaisAM16} combined CNN-based border detection with an optimized region growing and merging framework to achieve high average accuracy for dish segmentation. It was improved from the previous work \citep{DBLP:conf/iciap/DehaisAM15}, and yet the computational resources were costly. Following the GoCARB system, the team developed $\text{goFOOD}^\text{TM}$ \citep{goFOODTM}. For semi-automatic segmentation, region growing and merging algorithms were still used. They also developed an automatic food segmentation method based on the Mask-RCNN \citep{he2017mask}. For the recognition module, an improved Inception V3 was used to achieve hierarchical food recognition. GoCARB focused on calculating carbohydrates, while $\text{goFOOD}^\text{TM}$ can be used to estimate the calories and nutrients in a meal. These methods are easy to implement. However, they are limited by the complexity of the food scene (e.g., background, lighting, inter-class variability of food). Recently, they were committed to studying the nutritional intake of hospitalized patients \citep{DBLP:journals/corr/abs-2003-08273}. Through the development of a series of novel AI algorithms, the goal of achieving high performance on a small amount of training data has been achieved. \cite{DBLP:journals/corr/abs-2003-08273} proposed the multi-task contextual network to simultaneously segmented plates and foods. In addition, the application of the contextual layer enhanced the relationship between food and plate types and thus resulted in improved segmentation accuracy. Due to the finite nature of food samples, they trained one few-shot learning-based classifier for food recognition in the meta-learning framework \citep{DBLP:conf/nips/SnellSZ17}. The core idea was to learn the transferred knowledge in many similar tasks and use it for new tasks. It was a fully automated pipeline system based on AI, which promoted further development in nutrient assessment.
\lth{In addition to these works, \cite{DBLP:journals/mlc/AslanCMS20} benchmarked many deep learning-based food segmentation methods on their proposed dataset Food50Seg. Food/no-food segmentation and semantic segmentation methods were benchmarked from many different perspectives, which will guide the design and improvement of food analysis systems. }

\subsection{Food volume estimation}
A common approach to estimate food quality is to evaluate its volume, which can be used in conjunction with a food density database to gauge the quality of food in the image. Therefore, this section mainly focuses on volume estimation. Area and depth estimation based on top and side views can be directly used to estimate the food volume \citep{DBLP:journals/tim/PouladzadehSA14}. However, this method requires the user's thumb to be a reference object and is not accurate. Therefore, the 3D reconstruction methods are commonly used to estimate the volume. It relies on computer vision techniques to add missing dimensions from 2D images to create a 3D space, using visual cues to understand the third dimension of objects.

3D reconstruction starts with the calibration of camera parameters to perform volume reconstruction. Most methods use fiducial markers to facilitate the camera calibration process. Traditionally, spherical objects \citep{DBLP:journals/pami/ZhangWZ07} and the checkerboard patterns \citep{yu2006robust} are used as fiducial markers. 3D reconstruction includes methods that use a single view or multiple views as the input. Reconstruction based on a single view is an ill-posed problem, but the geometric model is conducive to recover the 3D parameters of the food in the scene. \cite{DBLP:journals/jstsp/ZhuBWKBED10} used both the spherical approximation model and prismatic approximation model for 3D volume reconstruction. Extracting feature points from food images based on segmentation, constructing base surfaces for 3D volume estimation. \cite{chen2013model} constructed a 3D geometric shape model library, including regular and irregular models. The food contours are matched to the 3D shape model, and the food volume is estimated according to the size of the model. However, the shape model library requires heavy user input, and the results are unstable. There are some studies, which utilize the prior knowledge of \enquote{container shape} as the geometric context information to obtain food volume \citep{DBLP:conf/mm/XuHKPBD13, DBLP:conf/ism/FangLZDB15}. In comparison, the method proposed by \cite{8329671} does not require prior knowledge such as shape models. They adopted the simultaneous localization and mapping system based on the monocular vision for dynamic food volume measurement. The optimized multi-convex hull algorithm was applied to the sparse map to form a 3D mesh object and continuously measured the food volume.

Based on multiple views, sparse reconstruction or dense reconstruction can be performed. The 3D models reconstructed using sparse feature points have lower resolution \citep{DBLP:journals/percom/KongT12}. In dense reconstruction, the stereo matching method is most commonly used. \cite{DBLP:conf/wacv/PuriZYDS09} composed two food images into a stereo pair. According to the hierarchal pyramid matching scheme, the left and right images were matched to perform the 3D reconstruction. Based on the reconstructed 3D point cloud, the scale was further estimated, and the 3D volume of the food was calculated. This method takes full advantage of dense reconstruction. However,  the whole process is labor-intensive. \cite{DBLP:conf/sitis/RahmanLPFKBD12} proposed a volume estimation method based on a set of images taken from the left and right sides of the food. They converted the reconstructed 3D point cloud into a series of slices and added the volume of each slice to obtain the total volume of the food. Although many attempts on volume estimation were made, most of them relied on user input as well as strong assumptions. \cite{dehais2016two} proposed a dense two-view reconstruction method with a reference card. The method made minimal assumptions about the type or shape of the food and was highly automated. They added disparity range extraction to the stereo matching process. A disparity map is a depth map where the depth information is derived from offset images of the same scene. Once the disparity map is estimated, the disparities can be converted to depth according to the following:

\begin{equation}
     \text{depth}=\dfrac{Bf}{\text{disparity}}\label{disparity}
\end{equation}
where  $B$ is the length of the baseline connecting the camera centers and $f$ is the focal length of the camera. Using the depth map and camera parameters, the 3D positions of all the points in the food image can be restored. In addition to the depth information from stereo-matched images, depth information can be easily obtained using a depth camera. \cite{DBLP:conf/siggrapha/ChenYHWLCYO12} introduced a quality estimation method using the depth information. Take the \enquote{sour and spicy soup} as an example. They used a depth camera Kinect to obtain the color and depth information of the noodle soup. The area of the bowl and the depth of the food is used for quantity estimation. However, the Kinect is a depth camera of structured infrared light, which affects the depth estimation due to the light reflection. \cite{DBLP:conf/iccv/MyersJRKGSGPH015} adopted depth camera-based 3D volume estimation. The depth map was converted to a voxel representation using CNN and random sample consensus (RANSAC) reconstruction. With the development of technology, stereo images can be quickly acquired through devices with stereo cameras such as 3D smartphones, and thus becomes another type of popular methods.

Later, researchers have been exploring more convenient and accurate ways to estimate the volume. For example, \cite{yang2019image} argued that carrying fiducial markers was very inconvenient and affected subsequent image processing. Therefore, they proposed a novel virtual reality-based volume estimation method. The key idea of eliminating fiducial markers was to determine the camera orientation through the motion sensor inside the smartphone and determine the location of visible points based on the width or length of the phone. \cite{DBLP:journals/tii/LoSQL20} combined the strength of deep learning and 3D reconstruction and developed a point completion network to solve the problems of scale ambiguity and food occlusion.

\begin{figure}[htbp]
    \centering
    \includegraphics[width=\textwidth]{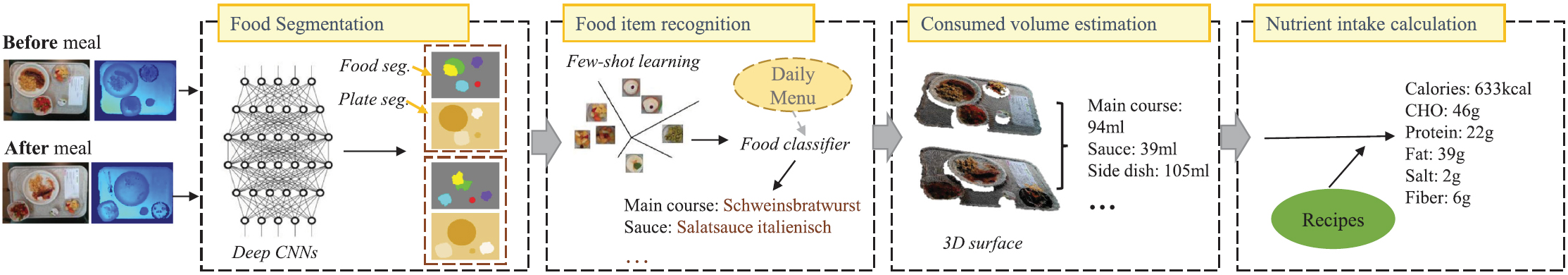}
    \caption{The framework of multi-stage dietary assessment \citep{DBLP:journals/corr/abs-2003-08273}.}
    \label{work:multi-stage}

\end{figure}

As one typical multi-stage VBDA work in Figure \ref{work:multi-stage}, in the image analysis phase, \cite{DBLP:journals/corr/abs-2003-08273} performed hyper food semantic segmentation and fine-grained food recognition. The next stage includes volume and weight estimation, as well as nutritional calculation. They used the Intel RealSense RGB-D Sensor to capture food images before and after a meal, which can output aligned RGB images and depth images simultaneously. Food consumption depends on the volume of food before and after a meal. To obtain the plate surface, the segmentation map allows to determine the plate's position and then enables the RANSAC algorithm to estimate the orientation of the plate. In addition, the acquired food depth image was converted into a point cloud. The triangular surfaces were divided by the Delaunay triangulation method for the construction of food 3D surfaces. They combined the plate surface estimation with 3D food surface extraction to further calculate the food volume. To calculate nutrition, they converted the volume to weight. According to the calculated weight and food category, the nutrient content of each food before and after meals was calculated. It can be seen that each stage has its specific task and the multi-stage architecture relies on the input and output of each stage.

\section{End-to-end approaches}
\label{E2EA}
The end-to-end approach focuses on using a single model to replace the pipeline of the multi-stage method. Considering that end-to-end methods have shown promising prospects in tasks such as image recognition \citep{DBLP:conf/cvpr/HeZRS16}, more and more end-to-end works for nutrition prediction are proposed.
\cite{fang2019end} proposed one novel end-to-end food energy estimation method, which used the generative adversarial network to estimate the image-to-energy mapping. The energy distribution map \citep{DBLP:conf/icip/FangSMFDZKB18} is a new method that replaces the \enquote{depth map} in \cite{DBLP:conf/iccv/MyersJRKGSGPH015} for visualizing the position of food in the images and how much relative energy was presented in different food regions.
\cite{DBLP:journals/EndtoEnd} improved this work \citep{DBLP:conf/icip/FangSMFDZKB18} by adding a localization network to detect food items in the image. Bounding boxes were applied to the energy distribution map to reduce the estimation errors.

However, all of these works only achieved an estimate of the calories in the food. In dietary assessment, multiple nutrient estimates (e.g., different types of macronutrients) are often required. There are also associated representations among other VBDA subtasks. Focusing on a single task tends to neglect information that might help  do better on the metrics we care about~\citep{ruder2017overview}.  By realizing the shared representation between related tasks, the method of learning multiple related tasks together is called Multi-Task Learning (MTL). Compared with a single task, it can improve the generalization performance of the task to realize a more comprehensive VBDA architecture. MTL is typically implemented with either soft or hard parameter sharing of hidden layers.

In soft parameter sharing, each task has its model with its parameters. The distance between the parameters of the model is then regularized to encourage the parameters to be similar.
\cite{DBLP:conf/mipr/HeSWKBZ20} performed food classification and portion size estimation in a multi-task framework with shared soft parameters. \lth{ They used the L2-norm to regularize the parameters of the two models. Finally, the output of both models is concatenated and then normalized to regress the calorie estimation result.}

Hard parameter sharing is generally applied by sharing the hidden layers among all tasks while reserving several task-specific output layers.
Ege and Yanai used this architecture in a series of end-to-end VBDA works.
\cite{DBLP:conf/mm/EgeY17} designed a multi-task CNN based on VGG-16 to extract features. After that, a Fully-Connected~(FC) layer is shared by all tasks. Another FC layer is branched to each task for the output unit, including predicting calories and food categories, ingredients, and cooking instructions. However, they assumed that food images only contain one dish. Their new work further used the improved YOLOv2 \citep{DBLP:conf/cvpr/RedmonDGF16} to apply object detection to multiple dishes \citep{DBLP:conf/ijcai/EgeY18}. The network can estimate the boundaries, categories, and calories of dishes from images of multiple dishes. However, the output of this method does not consider food amount and only corresponds to the calories of a dish. Considering there are no datasets on multiple-dish food images with both bounding boxes and food calories, two types of datasets were used alternately in \cite{DBLP:journals/ieicet/EgeY19} to train a single CNN.
The results showed that their multi-task approach achieved higher accuracy, higher speed, and smaller network size than a sequential model for food detection and calorie estimation.
Similarly, \cite{DBLP:journals/aai/SitujuTSYKL19} collected two image datasets at different scales for calorie and salinity assessment. They used the Xception \citep{Chollet_2017_CVPR} model to extract features. The network branches to each task from the global average-pooling layer. Each task has a global average pooling layer, a FC layer with the dropout, and an output layer, respectively. Two-stage fine-tuning was applied to improve the accuracy. The results proved that the multi-task CNN outperformed the single-task CNN.
\cite{DBLP:journals/corr/abs-2011-01082}  designed the network from a different perspective. They introduced a framework for retrieving nutritional information of recipes by matching ingredients and their mass to a nutrient database using phrase embedding. Pre-trained ResNet and DenseNet \citep{DBLP:conf/cvpr/HuangLMW17} architectures were leveraged as the backbone models. The last FC layer was replaced with a linear layer to regress the macronutrient information. A good distinction is made between high and low-calorie dishes.

The idea of multi-tasking learning had also been used to design some complex end-to-end VBDA networks.
\cite{DBLP:conf/ijcai/LuAASFM18} presented a MTL-based VBDA network, which simultaneously implemented food recognition, segmentation, and volume estimation. The feature extraction module is composed of ResNet50 and feature pyramid network. A Region Proposal Net (RPN) \citep{7485869} along with the feature extraction module preliminary produced the bounding box for each candidate object. MTL is applied to recognize food, regress bounding-box, and predict binary mask from the output of RPN. In terms of accuracy, the performance of food segmentation and volume estimation outperformed the state-of-the-art at that time. \cite{DBLP:journals/corr/abs-2008-00818} focused on reconstructing the 3D food model from a single RGB-image input without the use of depth information, and for the first time realized the entire pipeline single view diet assessment. They utilized the ResNet50 as the feature extraction module. The encoded features were decoded in several different branches to predict depth maps, semantic maps, and the camera pose between the consecutive frames.

\begin{figure}[htbp]
    \centering
    \includegraphics[width=\textwidth]{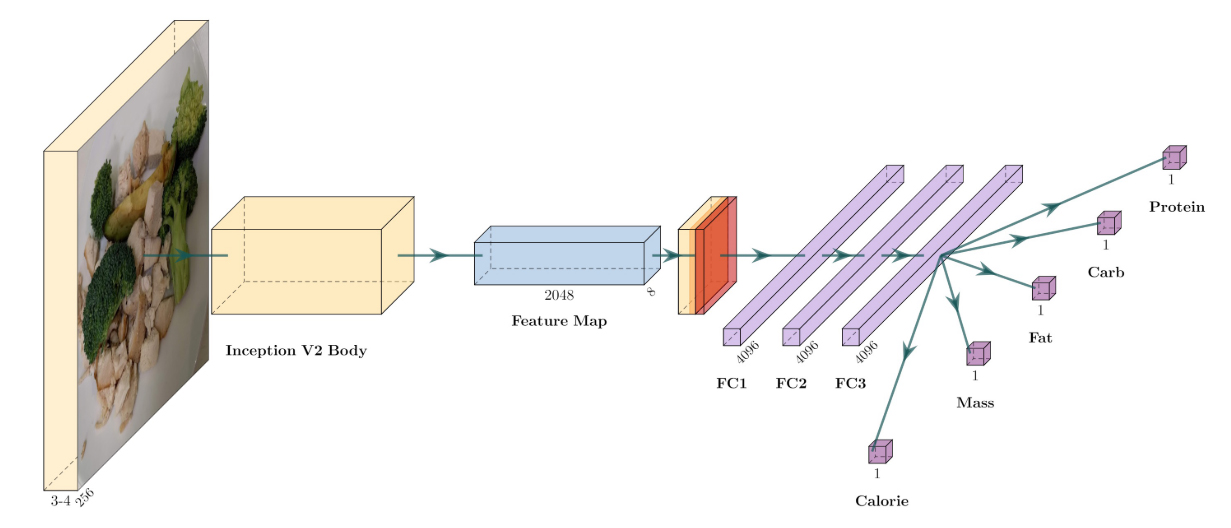}
    \caption{The framework of MTL end-to-end dietary assessment \citep{DBLP:journals/corr/abs-2103-03375}.}
    \label{work:end-to-end}

\end{figure}

\cite{DBLP:journals/corr/abs-2103-03375} carried out a series of VBDA experiments on their proposed Nutrition5k dataset, which covered the typical end-to-end works. The experiments was based on a MTL end-to-end network in Fig \ref{work:end-to-end}. The Inception V2 backbone encoder was used to extract feature maps. Then average-pooling was applied, and two FC layers followed. They trained a separate multi-task head for each regression task (calorie, macronutrients, and optionally mass), including two FC layers of different sizes. The experiments included estimating nutrients per gram from RGB image, estimating nutrients directly from RGB image, estimating nutrients directly from RGB-D image, and estimating nutrients from RGB image and portion. 
The results showed that the complexity of VBDA substantially increased when portion estimation is required, and depth information is essential for estimating nutrients when considering food portion size. Besides, the network outperformed professional nutritionists at caloric and macronutrient estimation in the generic food setting.

\section{Datasets and evaluation metrics}
\label{Datasets}
\subsection{Datasets}
In this section, we briefly summarize and compare various datasets for dietary assessment. Table \ref{datasettable} details the major nutrition-related datasets from different aspects, such as classes, total size of images, volume, calorie type and macroronutrients~(i.e., carbohydrate, protein and fat).

Most of the datasets related to dietary assessment consist of specific types of foods. For example, Menu-Match \citep{DBLP:conf/wacv/BeijbomJMSK15} focuses on the restaurant scenarios, and contains food images and associated nutritional information. FoodLog \citep{miyazaki2011image} contains user-uploaded food images, text, and calorie information. It is mainly composed of Japanese food. \cite{DBLP:conf/siggrapha/ChenYHWLCYO12} collected 50 kinds of Chinese food for food identification and quantity estimation. However, only food recognition and quantity estimation was performed and no nutritional information was labeled. Later, \cite{ma2021image} built the first Chinese food image dataset ChinaFood-100 with comprehensive descriptions and nutritional information annotations (i.e., protein, fiber, vitamin C, calcium and iron). Recently, Nutrition5k \citep{DBLP:journals/corr/abs-2103-03375} is released for the development of dietary assessment. It consists of food dishes with corresponding video streams, ingredient weights, depth images and high-precision nutrition annotations (i.e., carbohydrate, protein and fat).

As we all know, the depth image of food is beneficial for volume estimation. \cite{DBLP:journals/corr/abs-2003-08273} used the Intel RealSense RGB-D sensor to acquire both RGB images and depth images of food, and constructed the NIAD dataset. It should be emphasized that there are few datasets for food volume estimation. The annotations needed in such datasets are pixel-level food segmentation maps, image-level labels, and basic facts about food volume. MADiMa \citep{DBLP:conf/iciap/AllegraADLSFM17} is a dataset that meets these requirements. It contains RGB images and depth maps of meals, segmentation and recognition maps, 3D models, weights and volumes. Therefore, many studies have conducted experiments on this dataset to provide baseline performance of food segmentation, depth, and volume estimation. For example, \cite{goFOODTM} evaluated the performance of the entire process of dietary assessment based on this MADiMa dataset. Considering that  MADiMa only contains meal images taken by a monocular camera, they created one fast food dataset from McDonald's to support two-view images and stereo image pairs as input to comprehensively evaluate the performance of $\text{goFOOD}^\text{TM}$. \cite{DBLP:conf/ijcai/LuAASFM18} also used the MADiMa dataset as the training and evaluation dataset. \cite{DBLP:journals/corr/abs-2008-00818} used it to evaluate the performance of the proposed method. Furthermore, the Canteen dataset \citep{DBLP:journals/corr/abs-2008-00818} was proposed as a real scenario-based evaluation compared with MADiMa, which was created in a laboratory environment.

In addition, many studies have proposed the dataset of eating occasion images for food energy datasets \citep{fang2019end, DBLP:conf/mipr/HeSWKBZ20, DBLP:journals/EndtoEnd}. The distribution of food energy in the eating scene can be represented as an energy distribution image. The purpose is to estimate the energy values of the images based on the estimated energy distribution. Such datasets usually contain annotations of food categories, bounding box information, and the food size. To verify the effectiveness of multi-task learning, \cite{DBLP:conf/mm/EgeY17} constructed two datasets of Japanese and American with food calories, categories, ingredients and cooking directions. In addition, due to the high cost of manually creating nutrition annotations, they chose to collect data from the web. Many researches were done using this approach. For example, \cite{DBLP:journals/corr/abs-2011-01082} collected the dataset from a German recipe website, namely the pic2kcal dataset, which contains cooking instructions, ingredients and images. \cite{DBLP:journals/aai/SitujuTSYKL19} collected large datasets of food images with calorie and salinity annotations based on cooking food websites. 
The proliferation of these recipe-sharing websites has facilitated the collection of food datasets. For example, Ajinomoto Park includes food ingredients, cooking instructions and rich nutritional information. Allrecipes also contains the above information and adds a user comment module. They are usually provided by experts such as culinary researchers or cooks and thus has excellent credibility. With the rapid development of technological infrastructure, web-based data collection methods will become more popular.


    %

\newcolumntype{P}[1]{>{\Centering\hspace{0pt}}p{#1}}
\newcolumntype{F}[1]{>{\hspace{0pt}}p{#1}}
\newcolumntype{Z}{>{\centering\let\newline\\\arraybackslash\hspace{0pt}}X}
\begin{table}[H]
\renewcommand\arraystretch{1.5}

    \caption{Nutrition-related datasets.}\label{datasettable}
   \scriptsize
    \begin{tabularx}{\textwidth}{F{2.8cm}F{2cm}XF{3.2CM}}
    \toprule
     Dataset&Total\# images/class&Data type&References\\
    \midrule
       
        \textbf{Multi-stage}&&&\\
         FoodLog&6,512(2,000)&Images with calorie values&\cite{miyazaki2011image}\\
        

        FOOD&676(49)&Images with categories, weight and nutrient composition&\cite{DBLP:conf/aaaifs/EskinM12}\\
        
        
        Menu-Match &646(41)&Images with tags and calorie values&\cite{DBLP:conf/wacv/BeijbomJMSK15}\\
        
        
        FooDD&3k(30)&Images with categories&\cite{DBLP:conf/iciap/PouladzadehYS15}\\
        
        \lth{Inselspital dataset}&  \lth{1620(248)} & \lth{Images with categories, bounding boxes, actual volume and CHO grams} & \cite{anthimopoulos2015computer}\\
        Okamoto&120(20)&Images with calorie values& \cite{DBLP:conf/mm/OkamotoY16}\\
        
        Chokr&1,132(5)&Images with categories, size and calorie values&\cite{DBLP:conf/aaai/ChokrE17}\\
        
        MADiMa&21,807(80)&Images with depth maps, weight, volume and nutrient composition&\cite{DBLP:conf/iciap/AllegraADLSFM17}\\



        
        
        Fast food&20(14)&Image pairs with categories and nutrient composition&\cite{goFOODTM}\\
        
        
        VIPER-FoodNet dataset&14,991(82)&Images with categories and bounding boxes&\cite{mao2021visual}\\
         
       
        ChinaFood-100&10,047(100)&Images with nutrient composition&\cite{ma2021image}\\

        NIAD&1,281(521)&Images with depth, recipes and nutrient composition&\cite{DBLP:journals/corr/abs-2003-08273}\\
        
        \ww{ChinaMartFood-109}&\ww{10,921(18)}&\ww{Images with nutrient composition}&\cite{ma2022application}\\
        
       \midrule
       \textbf{End-to-end}&&&\\
        American recipe dataset&2,848(21)&Images with categories and recipe& \multirow{2}*{\cite{DBLP:conf/mm/EgeY17}}\\
        Japanese recipe dataset&4,877(15)&Images with categories and recipe&\\
        
        
        Situju&3,051(14)&Images with calories and salinity values& \cite{DBLP:journals/aai/SitujuTSYKL19}\\
        
        Image-to-Energy dataset&-(79)&Images with categories, segmentation masks and calorie values&\cite{fang2019end}\\

        Eating occasion image to food energy dataset&834(21)&Images with categories and portion size groundtruth&\cite{DBLP:conf/mipr/HeSWKBZ20}\\
        
        Eating occasion image to food energy dataset&915(31)&Images with bounding box, categories and calorie values&\cite{DBLP:journals/EndtoEnd}\\
        
        pic2kcal&308,000(70,000)&Images with categories, recipe and nutrient composition&\cite{DBLP:journals/corr/abs-2011-01082}\\
                
        Nutrition5k&20k(5k)& Multimedia data includes video streams and depth images with component weights, and nutrient composition&\cite{DBLP:journals/corr/abs-2103-03375}\\

        Canteen dataset&16k(80)&Video frames with annotation&\cite{DBLP:journals/corr/abs-2008-00818}\\

        \bottomrule
    \end{tabularx}
\end{table}

\subsection{Evaluation metrics}
One dietary assessment system should be evaluated in multiple respects, such as accuracy, run time, and resource requirements.
However, most researchers have only focused on metrics for assessment accuracy. For the multi-stage method, the evaluation should be performed in each stage. In contrast, the end-to-end method should be evaluated as a whole, which is similar to the final stage of the multi-stage method. Due to the difference of each stage, evaluation metrics are classified and summarized by the output type.

For food image analysis, Top-1 Accuracy and Top-5 Accuracy are often used to evaluate the performance of recognition. 
Here, Top-N accuracy means the probability that the correct class of food  presents in Top-n recognition results.
Detection procedures will mark out targets with bounding boxes. The area in the bounding box is a set of pixels. Segmentation procedures will output a mask of the input image. For one particular category in the image, its mask is also a set of pixels. As a result, the evaluation metrics of food detection and segmentation are similar, and they are evaluated using the Intersection over Union (IoU):

\begin{equation}
    \text{IoU} = J(A,B) = \dfrac{\left| A\cap B\right|}{\left| A\cup B\right|}
\end{equation}
where IoU is defined as the area of intersection between the predicted map $A$ and the ground truth map $B$, divided by the area of the union between the two maps, and ranges between 0 and 1. Mean-IoU (mIoU) is further defined as the average IoU over all classes \citep{minaee2021image}. Precision / Recall / F1 score can be defined for each class, as well as at the aggregate level, as follows:

\begin{equation}
    \text{Precision} = \dfrac{\text{TP}}{\text{TP}+\text{FP}}\\
    \text{Recall} = \dfrac{\text{TP}}{\text{TP}+\text{FN}}
\end{equation}
where TP, FP, FN refer to the true positive fraction, the false positive fraction, and  the false negative fraction, respectively.
\lth{The Precision-Recall curve shows the tradeoff between the Precision and Recall values for different thresholds. This curve helps to select the best threshold to maximize both metrics.
Compared to graphically deciding the best values for Precision and Recall, using the F1 score to measure their balance is better. It is defined as the harmonic mean of Precision and Recall:}

\begin{equation}
    \text{F1} = \dfrac{2\ \text{Precision Recall}}{ \text{Precision}+\text{Recall}}
\end{equation}
\lth{When the value of F1 is high, this means both the Precision and Recall are high. In addition, the Average Precision (AP) is a way to summarize the Precision-Recall curve into a single value representing the average of all Precisions \citep{mAP}. It is defined as the weighted sum of Precisions at each threshold where the weight is the increase in Recall. The mean of the APs for all classes is the mAP.}

\lth{Besides, Pixel Accuracy (PA) is also widely used in evaluating segmentation. Its definition is the ratio of correctly classified pixels divided by the total number.
\begin{equation}
    \text{PA} = \dfrac{\text{The number of correctly classified pixels}}{ \text{The total number of pixels}}
\end{equation}}

Considering the output type of portion estimation, nutrient estimation, and end-to-end methods is numeric, Mean Absolute Error (MAE) and Mean Absolute Percentage Error (MAPE) are usually used in the evaluation. 

\begin{equation}
        \text{MAE} = \dfrac{1}{N}\sum\limits_{i = 1}^N {\left| {\mathop {\tilde w}\nolimits_i  - \mathop {\bar w}\nolimits_i } \right|} 
\end{equation}

\begin{equation}
        \text{MAPE} = \dfrac{{1}}{N}\sum\limits_{i = 1}^N {\dfrac{{\left| {\mathop {\tilde w}\nolimits_i  - \mathop {\bar w}\nolimits_i } \right|}}{{\mathop {\bar w}\nolimits_i }}} 
\end{equation}
where $\mathop {\tilde w}\nolimits_i$ is the estimated portion size of the $i$-th image, $\mathop {\bar w}\nolimits_i$ is the groundtruth value of $i$-th image and $N$ is the number of testing images \citep{DBLP:journals/corr/abs-2103-07562}.

\section{Future works}
\label{FutureWorks}
According to comprehensive discussions on existing efforts, we now articulate key open challenges and give some prospective analysis on future works for VBDA.
\subsection{Large-scale benchmark dataset construction}
Currently, VBDA works are driven mainly by data and thus highly rely on the construction of datasets. However, most works tend to construct their own datasets for specific tasks rather than reusing existing datasets. Besides, accurate portion and nutrition annotation are extremely burdensome and costly to obtain, which limits the scale of VBDA datasets. These reasons lead to the lack of large-scale VBDA benchmarks.
There are two representative datasets, namely MADiMa \citep{DBLP:conf/iciap/AllegraADLSFM17} and recently released Nutrition5k \citep{DBLP:journals/corr/abs-2103-03375}.  The MADiMa dataset contains food images, depth maps, segmentation maps, weight/volume measurements of served meals, nutrient content together with the corresponding annotations, labels and accelerometer data. It is constructed for multi-stage VBDA.  Segmentation map plays a vital role in multi-stage VBDA. However, it is not necessary for an end-to-end approach. Nutrition5k \citep{DBLP:journals/corr/abs-2103-03375} thus drops the segmentation maps and adds food ingredient information for an end-to-end VBDA. Ingredient information gives a new perspective to understand nutrients and enables the model to handle unseen food categories. However, these two datasets only contain western food with small scale. The standard benchmark dataset should contain not only western cuisines, but also eastern ones, and cover a wide range of food types from all over the world. Different cuisines differ in many aspects, such as the 
level of separation of ingredients, and cooking procedures. The ingredients from western cuisines are generally separate while ones from eastern ones are often mixed.  The cooking procedure in the eastern cuisine is more various than western ones. Such differences make the VBDA solutions probably different. And for each item, there should be corresponding images, depth map, weight, nutrition information, recipe, and ingredient distribution map. 
Therefore, the VBDA dataset with various cuisine types should be constructed to enable targeted solutions for different cuisines, resulting in more  widespread use and popularization of VBDA.
Because of the data-driven nature of deep learning, a large-scale VBDA dataset is also urgently needed for developing advanced VBDA algorithms.

\subsection{Fine-grained visual analysis}

Compared with ordinary objects, many kinds of food do not display unique spatial configurations. They are generally non-rigid, and food appearance can be diverse depending on the used ingredients and the cooking styles. Additionally, hidden ingredients and distinctive cooking methods (e.g., stir-fry, steaming, boiling, roasting) present obstacles to assessing the nutrient composition of different dishes. 
In addition, external factors such as illumination are also a pressing issue in image analysis.
In order to solve these problems, we should implement more fine-grained visual analysis and present three possible future directions in detail.
First, more complex and advanced CNNs and Transformers have demonstrated powerful capabilities in a range of visually related tasks~\citep{khan2021transformers}. Similarly, we believe that a new network structure designed for VBDA allows for better extraction of visual features and fine-grained visual analysis.
Second, the performance of visual analysis can be improved using pre-trained models on the large-scale food analysis dataset~\citep{Weiqing-Food500-MM2020,Weiqing-LSVFR-CoRR2021}. For example, \cite{Weiqing-LSVFR-CoRR2021} verified the better generalization of their proposed large-scale dataset Food2K in various vision tasks, such as food recognition, detection, and segmentation.  We can study transfer learning from more aspects in the future, such as cross-cuisine, cross-scenario, and cross super-class. Transfer learning expands the training data in a sense which makes the model more robust and enhances the generalization ability.
Third, besides improving the performance of neural networks, utilizing food recipes can also help realize fine-grained visual analysis. For example,
\cite{10.1145/3347448.3357163} suggested that multimodal recipe retrieval could be the first step in a pipeline for nutrient assessment. Understanding the food composition is more scalable than identifying food categories. The ingredient information and cooking methods in food recipes can establish the nutritional connection among relative foods and help the model understand different visual forms of the same ingredient.

\subsection{Accurate food volume estimation}
Volume estimation is a commonly used method of portion estimation. However, the performance of current volume estimation portion is not yet satisfactory. Existing studies mainly focused on model-based or stereo-based methods. The first food volume estimation system constructed 3D geometric shape model libraries \citep{chen2013model}. They matched 2D food contours to the 3D shape model, and estimated the food volume based on the size of the model. However, the shape model library is not applicable to all food objects. To solve the problem, multi-view based approaches are developed to perceive depth according to more spatial information of food geometry \citep{DBLP:conf/wacv/PuriZYDS09, DBLP:journals/percom/KongT12, dehais2016two}. However, it is inconvenient for users to take multiple food images from different angles. Although food volume estimation methods based on a single view are generally preferred, such methods cannot provide complete information about the 3D shape of the food (e.g., the back of the food is not visible). Later, the depth information of food images is explored to  facilitate the food localization and 3D reconstruction. However, the depth images estimated based on food images are not accurate enough. Therefore, \cite{DBLP:journals/corr/abs-2003-08273} simultaneously acquired RGB and depth images of food using the RealSense RGB-D sensor. They converted the acquired depth images into the point cloud to further calculate the food volume. If more mobile devices with depth images acquisition sensors could be developed and popularized, it would simplify the volume estimation to a great extent. Additionally, \cite{DBLP:journals/corr/abs-2104-00877} found that the spatial structure information in the scene plays a crucial role in depth perception. They proposed S2R-DepthNet to remove the influence on the texture information and utilized the human visual system features for spatial structure grasping to enable the network the powerful generalization capability for depth estimation. This is because the human visual system can perceive depth through stereo vision of both eyes. Based on the accumulation of many natural scenes, humans can judge the distance between objects in the image when there is only one image. Therefore, integrating human visual observation features of structural information into volume estimation is probably one future direction.

\subsection{Practical VBDA solution}
There has been considerable effort to achieve fully automated, highly accurate nutrient assessments in the academic field. 
However, the applications available to end-users still mainly rely on the manual recording of food types and amounts, indicating a dichotomy between the academic  and the industrial fields.
There is no doubt that the accuracy of visual analysis and volume estimation affects the accuracy of the results.
Besides, the scenario provides additional context relations and can further improve the assessment accuracy.
In a scenario-specific dietary assessment, making full use of the knowledge of the scenario is the key to improve usability. For example, in the canteen environment, the type of food is fixed, specific containers will be used to hold specific types of food.
In addition, specialized devices such as depth cameras can be deployed to gather more information. \cite{DBLP:journals/corr/abs-2103-03375}  proved the depth information improved the accuracy of nutrient assessment in the control experiment. 
However, it is not practical to require users to acquire such equipment as depth cameras. Considering  smartphones are still the most widely used portable device in daily dietary assessments. We can also obtain rich multimodal information from the smartphone’s camera, positioning system, and inertial measurement sensors. Therefore, designing solutions specifically for portable devices, such as smartphones, is a hot topic for research. However, the solutions are confined to a laboratory environment and have many limitations in practice, such as the scarcity of analyzable food types and the lack of large-scale opening testing.
In addition, complex eastern cuisines are more difficult so that existing VBDA methods can not obtain satisfactory assessment performance.  Therefore, towards realizing a practical VBDA solution, there is still a long way to go.

\subsection{Intelligent dietary management}
People generally understand the connection between diet and health. However, the concept cannot be applied to practice, and it is difficult for them to reasonably control their dietary intake in daily life. For example, excessive sodium intake over a long period can raise blood pressure and become a risk factor for hypertension. Therefore, dietary management can help people solve potential dietary problems and achieve personalized health needs. Intelligent food record and food recommendation are two critical aspects of intelligent dietary management. 

For the intelligent food record, FoodLog \citep{kitamura2008food} has been created to process images taken by users. We expect to conduct statistics on the practical and diversified food data spontaneously generated by users. Food nutrition information is obtained and recorded through VBDA, generating a weekly or monthly diet record sheet for users. By analyzing the diet record sheet, we can decide that the intake of a certain type of food should be reduced or increased, thus assisting users in dietary management. For the food recommendation, intelligent food record can obtain the user’s real-time eating habit information, and then learn the dynamics of user preferences based on VBDA~\citep{delarue2004dynamics}, and finally combine user's health information with the learned user preference to realize personalized dietary recommendations~\citep{8930090}. Furthermore, VBDA-based recommendations can be used as an auxiliary tool for nutritionists. For example, with the popularity of the healthy and green living concept, more and more people prefer healthy weight loss. It is undeniable that healthy weight loss must not only ensure a balanced diet, but also meet the nutritional needs of the body and exercise. Based on the dietary recommendations provided by VBDA, nutritionists with rich professional knowledge further provide reasonable nutritional requirements for the body and exercise. In summary, VBDA has obvious potential and advantages in automatically monitoring diet and compliance with nutritional interventions for intelligent dietary management.

\section{Conclusions}
\label{Conclusion}
VBDA is currently a hot and challenging research field, which has gradually replaced traditional dietary assessment methods. In this review, we divide existing VBDA methods into two types of architectures. One  is a multi-stage VBDA architecture, which mainly consists of three parts: food image analysis, portion estimation, and nutrient derivation. Each stage has its own specific task and is linked to each other for nourishment. The end-to-end VBDA architecture emphasizes specifying the original input and nutritional output without multiple steps. The food datasets and evaluation metrics related to VBDA have also been clarified in the article. Although VBDA has made good progress, it has great potential to further enhance the practicality of VBDA. For example, it is urgent to build a large-scale benchmark dataset to enable the faster development of VBDA. How to achieve fine-grained visual analysis and accurate volume estimation  based on food characteristics are worthy of further exploration. The promising directions for VBDA are discussed in this paper, and we look forward to seeing a wide range of VBDA applications in the future.

\section*{Declaration of competing interest}
There is no conflict of interest in this article.

\section*{Acknowledgement}




This work was supported by the Scientific Research Program of Beijing Municipal Education Commission (grant No. KZ202110011017),
National Natural Science Foundation of China (grant No.
61877002, 61972378, U1936203, U19B2040) and Open Research Fund
of Beijing Key Laboratory of Big Data Technology for Food Safety (grant
No. BTBD-2020KF04)




\bibliographystyle{model5-names-remake}

\bibliography{cas-refs}

\begin{thebibliography}{128}
\expandafter\ifx\csname natexlab\endcsname\relax\def\natexlab#1{#1}\fi
\providecommand{\url}[1]{\texttt{#1}}
\providecommand{\href}[2]{#2}
\providecommand{\path}[1]{#1}
\providecommand{\DOIprefix}{}
\providecommand{\ArXivprefix}{arXiv:}
\providecommand{\URLprefix}{}
\providecommand{\Pubmedprefix}{pmid:}
\providecommand{\doi}[1]{\href{#1}{\path{#1}}}
\providecommand{\Pubmed}[1]{\href{pmid:#1}{\path{#1}}}
\providecommand{\bibinfo}[2]{#2}
\ifx\xfnm\relax \def\xfnm[#1]{\unskip,\space#1}\fi
\bibitem[{{Ahmed Fawzy Gad}(2021)}]{mAP}
\bibinfo{author}{{Ahmed Fawzy Gad}} (\bibinfo{year}{2021}).
\newblock \bibinfo{title}{{Evaluating Object Detection Models Using Mean
  Average Precision (mAP)}}.
\newblock \URLprefix \url{https://blog.paperspace.com/mean-average-precision}
  \bibinfo{note}{. {Accessed} January 13, 2022}.
\bibitem[{Allegra et~al.(2017)Allegra, Anthimopoulos, Dehais, Lu, Stanco,
  Farinella \& Mougiakakou}]{DBLP:conf/iciap/AllegraADLSFM17}
\bibinfo{author}{Allegra, D.}, \bibinfo{author}{Anthimopoulos, M.},
  \bibinfo{author}{Dehais, J.}, \bibinfo{author}{Lu, Y.},
  \bibinfo{author}{Stanco, F.}, \bibinfo{author}{Farinella, G.~M.}, \&
  \bibinfo{author}{Mougiakakou, S.} (\bibinfo{year}{2017}).
\newblock \bibinfo{title}{A multimedia database for automatic meal assessment
  systems}.
\newblock In {\it \bibinfo{booktitle}{International Conference on Image
  Analysis and Processing}\/} (pp. \bibinfo{pages}{471--478}).
\newblock \DOIprefix\doi{https://doi.org/10.1007/978-3-319-70742-6\_46}.
\bibitem[{Ando et~al.(2019)Ando, Ege, Cho \& Yanai}]{DBLP:conf/mm/AndoECY19}
\bibinfo{author}{Ando, Y.}, \bibinfo{author}{Ege, T.}, \bibinfo{author}{Cho,
  J.}, \& \bibinfo{author}{Yanai, K.} (\bibinfo{year}{2019}).
\newblock \bibinfo{title}{Depthcaloriecam: A mobile application for
  volume-based food calorie estimation using depth cameras}.
\newblock In {\it \bibinfo{booktitle}{Proceedings of the 5th International
  Workshop on Multimedia Assisted Dietary Management}\/} (pp.
  \bibinfo{pages}{76--81}).
\newblock \DOIprefix\doi{https://doi.org/10.1145/3347448.3357172}.
\bibitem[{Anthimopoulos et~al.(2015)Anthimopoulos, Dehais, Shevchik, Ransford,
  Duke, Diem \& Mougiakakou}]{anthimopoulos2015computer}
\bibinfo{author}{Anthimopoulos, M.}, \bibinfo{author}{Dehais, J.},
  \bibinfo{author}{Shevchik, S.}, \bibinfo{author}{Ransford, B.~H.},
  \bibinfo{author}{Duke, D.}, \bibinfo{author}{Diem, P.}, \&
  \bibinfo{author}{Mougiakakou, S.} (\bibinfo{year}{2015}).
\newblock \bibinfo{title}{Computer vision-based carbohydrate estimation for
  type 1 patients with diabetes using smartphones}.
\newblock {\it \bibinfo{journal}{Journal of Diabetes Science and
  Technology}\/},  {\it \bibinfo{volume}{9}\/}, \bibinfo{pages}{507--515}
  \DOIprefix\doi{https://doi.org/10.1177/1932296815580159}.
\bibitem[{Aslan et~al.(2020)Aslan, Ciocca, Mazzini \&
  Schettini}]{DBLP:journals/mlc/AslanCMS20}
\bibinfo{author}{Aslan, S.}, \bibinfo{author}{Ciocca, G.},
  \bibinfo{author}{Mazzini, D.}, \& \bibinfo{author}{Schettini, R.}
  (\bibinfo{year}{2020}).
\newblock \bibinfo{title}{Benchmarking algorithms for food localization and
  semantic segmentation}.
\newblock {\it \bibinfo{journal}{International Journal of Machine Learning and
  Cybernetics}\/},  {\it \bibinfo{volume}{11}\/}, \bibinfo{pages}{2827--2847}
  \DOIprefix\doi{https://doi.org/10.1007/s13042-020-01153-z}.
\bibitem[{Ayon et~al.(2021)Ayon, Mashrafi, Yousuf, Hossain \&
  Hossain}]{ayon2021foodiecal}
\bibinfo{author}{Ayon, S.~A.}, \bibinfo{author}{Mashrafi, C.~Z.},
  \bibinfo{author}{Yousuf, A.~B.}, \bibinfo{author}{Hossain, F.}, \&
  \bibinfo{author}{Hossain, M.~I.} (\bibinfo{year}{2021}).
\newblock \bibinfo{title}{{FoodieCal}: A convolutional neural network based
  food detection and calorie estimation system}.
\newblock In {\it \bibinfo{booktitle}{National Computing Colleges
  Conference}\/} (pp. \bibinfo{pages}{1--6}).
\newblock \DOIprefix\doi{https://doi.org/10.1109/NCCC49330.2021.9428820}.
\bibitem[{Beijbom et~al.(2015)Beijbom, Joshi, Morris, Saponas \&
  Khullar}]{DBLP:conf/wacv/BeijbomJMSK15}
\bibinfo{author}{Beijbom, O.}, \bibinfo{author}{Joshi, N.},
  \bibinfo{author}{Morris, D.}, \bibinfo{author}{Saponas, T.~S.}, \&
  \bibinfo{author}{Khullar, S.} (\bibinfo{year}{2015}).
\newblock \bibinfo{title}{{Menu-Match}: Restaurant-specific food logging from
  images}.
\newblock In {\it \bibinfo{booktitle}{{IEEE} Winter Conference on Applications
  of Computer Vision}\/} (pp. \bibinfo{pages}{844--851}).
\newblock \DOIprefix\doi{https://doi.org/10.1109/WACV.2015.117}.
\bibitem[{Cai et~al.(2020)Cai, Chen, Wang, Luo, Liu, Wu, He, Wang, Liu, Liu
  et~al.}]{cai2020obesity}
\bibinfo{author}{Cai, Q.}, \bibinfo{author}{Chen, F.}, \bibinfo{author}{Wang,
  T.}, \bibinfo{author}{Luo, F.}, \bibinfo{author}{Liu, X.},
  \bibinfo{author}{Wu, Q.}, \bibinfo{author}{He, Q.}, \bibinfo{author}{Wang,
  Z.}, \bibinfo{author}{Liu, Y.}, \bibinfo{author}{Liu, L.} et~al.
  (\bibinfo{year}{2020}).
\newblock \bibinfo{title}{Obesity and covid-19 severity in a designated
  hospital in shenzhen, china}.
\newblock {\it \bibinfo{journal}{Diabetes care}\/},  {\it
  \bibinfo{volume}{43}\/}, \bibinfo{pages}{1392--1398}
  \DOIprefix\doi{https://doi.org/10.2337/dc20-0576}.
\bibitem[{Cao et~al.(2021)Cao, Watt, Nowell, Quach, Simpson, De~Melo~Ferreira,
  Agarwal, Chu, Srivastava, Anderson et~al.}]{cao2021mesenteric}
\bibinfo{author}{Cao, E.}, \bibinfo{author}{Watt, M.~J.},
  \bibinfo{author}{Nowell, C.~J.}, \bibinfo{author}{Quach, T.},
  \bibinfo{author}{Simpson, J.~S.}, \bibinfo{author}{De~Melo~Ferreira, V.},
  \bibinfo{author}{Agarwal, S.}, \bibinfo{author}{Chu, H.},
  \bibinfo{author}{Srivastava, A.}, \bibinfo{author}{Anderson, D.} et~al.
  (\bibinfo{year}{2021}).
\newblock \bibinfo{title}{Mesenteric lymphatic dysfunction promotes insulin
  resistance and represents a potential treatment target in obesity}.
\newblock {\it \bibinfo{journal}{Nature Metabolism}\/},  {\it
  \bibinfo{volume}{3}\/}, \bibinfo{pages}{1175--1188}
  \DOIprefix\doi{https://doi.org/10.1038/s42255-021-00457-w}.
\bibitem[{Chen et~al.(2013)Chen, Jia, Yue, Li, Sun, Fernstrom \&
  Sun}]{chen2013model}
\bibinfo{author}{Chen, H.-C.}, \bibinfo{author}{Jia, W.}, \bibinfo{author}{Yue,
  Y.}, \bibinfo{author}{Li, Z.}, \bibinfo{author}{Sun, Y.-N.},
  \bibinfo{author}{Fernstrom, J.~D.}, \& \bibinfo{author}{Sun, M.}
  (\bibinfo{year}{2013}).
\newblock \bibinfo{title}{Model-based measurement of food portion size for
  image-based dietary assessment using {3D/2D} registration}.
\newblock {\it \bibinfo{journal}{Measurement Science and Technology}\/},  {\it
  \bibinfo{volume}{24}\/}, \bibinfo{pages}{105701}
  \DOIprefix\doi{https://doi.org/10.1088/0957-0233/24/10/105701}.
\bibitem[{Chen et~al.(2012)Chen, Yang, Ho, Wang, Liu, Chang, Yeh \&
  Ouhyoung}]{DBLP:conf/siggrapha/ChenYHWLCYO12}
\bibinfo{author}{Chen, M.-Y.}, \bibinfo{author}{Yang, Y.-H.},
  \bibinfo{author}{Ho, C.-J.}, \bibinfo{author}{Wang, S.-H.},
  \bibinfo{author}{Liu, S.-M.}, \bibinfo{author}{Chang, E.},
  \bibinfo{author}{Yeh, C.-H.}, \& \bibinfo{author}{Ouhyoung, M.}
  (\bibinfo{year}{2012}).
\newblock \bibinfo{title}{Automatic chinese food identification and quantity
  estimation}.
\newblock In {\it \bibinfo{booktitle}{SIGGRAPH Asia 2012 Technical Briefs}\/}
  (pp. \bibinfo{pages}{1--4}).
\newblock \DOIprefix\doi{https://doi.org/10.1145/2407746.2407775}.
\bibitem[{Chen et~al.(2021)Chen, Wang, Chen \&
  Zeng}]{DBLP:journals/corr/abs-2104-00877}
\bibinfo{author}{Chen, X.}, \bibinfo{author}{Wang, Y.}, \bibinfo{author}{Chen,
  X.}, \& \bibinfo{author}{Zeng, W.} (\bibinfo{year}{2021}).
\newblock \bibinfo{title}{{S2R-DepthNet: Learning a Generalizable
  Depth-Specific Structural Representation}}.
\newblock In {\it \bibinfo{booktitle}{Proceedings of the IEEE/CVF Conference on
  Computer Vision and Pattern Recognition}\/} (pp.
  \bibinfo{pages}{3034--3043}).
\newblock \URLprefix \url{https://arxiv.org/abs/2104.00877}.
\bibitem[{Chi et~al.(2008)Chi, Chen, Chu \& Lo}]{chi2008enabling}
\bibinfo{author}{Chi, P.-Y.~P.}, \bibinfo{author}{Chen, J.-H.},
  \bibinfo{author}{Chu, H.-H.}, \& \bibinfo{author}{Lo, J.-L.}
  (\bibinfo{year}{2008}).
\newblock \bibinfo{title}{Enabling calorie-aware cooking in a smart kitchen}.
\newblock In {\it \bibinfo{booktitle}{International Conference on Persuasive
  Technology}\/} (pp. \bibinfo{pages}{116--127}).
\newblock \DOIprefix\doi{https://doi.org/10.1007/978-3-540-68504-3_11}.
\bibitem[{Chokr \& Elbassuoni(2017)}]{DBLP:conf/aaai/ChokrE17}
\bibinfo{author}{Chokr, M.}, \& \bibinfo{author}{Elbassuoni, S.}
  (\bibinfo{year}{2017}).
\newblock \bibinfo{title}{Calories prediction from food images}.
\newblock In {\it \bibinfo{booktitle}{Proceedings of the Thirty-First {AAAI}
  Conference on Artificial Intelligence}\/} (pp. \bibinfo{pages}{4664--4669}).
\newblock \URLprefix
  \url{http://aaai.org/ocs/index.php/IAAI/IAAI17/paper/view/14204}.
\bibitem[{Chollet(2017)}]{Chollet_2017_CVPR}
\bibinfo{author}{Chollet, F.} (\bibinfo{year}{2017}).
\newblock \bibinfo{title}{Xception: Deep learning with depthwise separable
  convolutions}.
\newblock In {\it \bibinfo{booktitle}{Proceedings of the IEEE Conference on
  Computer Vision and Pattern Recognition}\/}.
\newblock \URLprefix
  \url{https://openaccess.thecvf.com/content_cvpr_2017/html/Chollet_Xception_Deep_Learning_CVPR_2017_paper.html}.
\bibitem[{Christodoulidis et~al.(2015)Christodoulidis, Anthimopoulos \&
  Mougiakakou}]{christodoulidis2015food}
\bibinfo{author}{Christodoulidis, S.}, \bibinfo{author}{Anthimopoulos, M.}, \&
  \bibinfo{author}{Mougiakakou, S.} (\bibinfo{year}{2015}).
\newblock \bibinfo{title}{Food recognition for dietary assessment using deep
  convolutional neural networks}.
\newblock In {\it \bibinfo{booktitle}{International Conference on Image
  Analysis and Processing}\/} (pp. \bibinfo{pages}{458--465}).
\newblock \DOIprefix\doi{https://doi.org/10.1007/978-3-319-23222-5\_56}.
\bibitem[{Chu et~al.(2021)Chu, Yu, Chen, Tian \& Zhai}]{chu2021dietary}
\bibinfo{author}{Chu, C.-Q.}, \bibinfo{author}{Yu, L.-l.},
  \bibinfo{author}{Chen, W.}, \bibinfo{author}{Tian, F.-W.}, \&
  \bibinfo{author}{Zhai, Q.-X.} (\bibinfo{year}{2021}).
\newblock \bibinfo{title}{Dietary patterns affect parkinson's disease via the
  microbiota-gut-brain axis}.
\newblock {\it \bibinfo{journal}{Trends in Food Science \& Technology}\/},
  {\it \bibinfo{volume}{116}\/}, \bibinfo{pages}{90--101}
  \DOIprefix\doi{https://doi.org/https://doi.org/10.1016/j.tifs.2021.07.004}.
\bibitem[{Ciocca et~al.(2015)Ciocca, Napoletano \&
  Schettini}]{DBLP:conf/iciap/CioccaNS15}
\bibinfo{author}{Ciocca, G.}, \bibinfo{author}{Napoletano, P.}, \&
  \bibinfo{author}{Schettini, R.} (\bibinfo{year}{2015}).
\newblock \bibinfo{title}{Food recognition and leftover estimation for daily
  diet monitoring}.
\newblock In {\it \bibinfo{booktitle}{International Conference on Image
  Analysis and Processing}\/} (pp. \bibinfo{pages}{334--341}).
\newblock \DOIprefix\doi{https://doi.org/10.1007/978-3-319-23222-5_41}.
\bibitem[{Dehais et~al.(2015)Dehais, Anthimopoulos \&
  Mougiakakou}]{DBLP:conf/iciap/DehaisAM15}
\bibinfo{author}{Dehais, J.}, \bibinfo{author}{Anthimopoulos, M.}, \&
  \bibinfo{author}{Mougiakakou, S.} (\bibinfo{year}{2015}).
\newblock \bibinfo{title}{Dish detection and segmentation for dietary
  assessment on smartphones}.
\newblock In {\it \bibinfo{booktitle}{International Conference on Image
  Analysis and Processing}\/} (pp. \bibinfo{pages}{433--440}).
\newblock \DOIprefix\doi{https://doi.org/10.1007/978-3-319-23222-5\_53}.
\bibitem[{Dehais et~al.(2016)Dehais, Anthimopoulos \&
  Mougiakakou}]{DBLP:conf/mm/DehaisAM16}
\bibinfo{author}{Dehais, J.}, \bibinfo{author}{Anthimopoulos, M.}, \&
  \bibinfo{author}{Mougiakakou, S.} (\bibinfo{year}{2016}).
\newblock \bibinfo{title}{Food image segmentation for dietary assessment}.
\newblock In {\it \bibinfo{booktitle}{Proceedings of the 2nd International
  Workshop on Multimedia Assisted Dietary Management}\/} (pp.
  \bibinfo{pages}{23--28}).
\newblock \DOIprefix\doi{https://doi.org/10.1145/2986035.2986047}.
\bibitem[{Dehais et~al.(2017)Dehais, Anthimopoulos, Shevchik \&
  Mougiakakou}]{dehais2016two}
\bibinfo{author}{Dehais, J.}, \bibinfo{author}{Anthimopoulos, M.},
  \bibinfo{author}{Shevchik, S.}, \& \bibinfo{author}{Mougiakakou, S.}
  (\bibinfo{year}{2017}).
\newblock \bibinfo{title}{Two-view 3d reconstruction for food volume
  estimation}.
\newblock {\it \bibinfo{journal}{IEEE Transactions on Multimedia}\/},  {\it
  \bibinfo{volume}{19}\/}, \bibinfo{pages}{1090--1099}
  \DOIprefix\doi{https://doi.org/10.1109/TMM.2016.2642792}.
\bibitem[{Delarue \& Loescher(2004)}]{delarue2004dynamics}
\bibinfo{author}{Delarue, J.}, \& \bibinfo{author}{Loescher, E.}
  (\bibinfo{year}{2004}).
\newblock \bibinfo{title}{Dynamics of food preferences: a case study with
  chewing gums}.
\newblock {\it \bibinfo{journal}{Food Quality and Preference}\/},  {\it
  \bibinfo{volume}{15}\/}, \bibinfo{pages}{771--779}
  \DOIprefix\doi{https://doi.org/10.1016/j.foodqual.2003.11.005}.
\bibitem[{Ege et~al.(2019)Ege, Shimoda \& Yanai}]{DBLP:conf/mm/EgeSY19}
\bibinfo{author}{Ege, T.}, \bibinfo{author}{Shimoda, W.}, \&
  \bibinfo{author}{Yanai, K.} (\bibinfo{year}{2019}).
\newblock \bibinfo{title}{A new large-scale food image segmentation dataset and
  its application to food calorie estimation based on grains of rice}.
\newblock In {\it \bibinfo{booktitle}{Proceedings of the 5th International
  Workshop on Multimedia Assisted Dietary Management}\/} (pp.
  \bibinfo{pages}{82--87}).
\newblock \DOIprefix\doi{https://doi.org/10.1145/3347448.3357162}.
\bibitem[{Ege \& Yanai(2017)}]{DBLP:conf/mm/EgeY17}
\bibinfo{author}{Ege, T.}, \& \bibinfo{author}{Yanai, K.}
  (\bibinfo{year}{2017}).
\newblock \bibinfo{title}{Image-based food calorie estimation using knowledge
  on food categories, ingredients and cooking directions}.
\newblock In {\it \bibinfo{booktitle}{Proceedings of the on Thematic Workshops
  of {ACM} Multimedia 2017}\/} (pp. \bibinfo{pages}{367--375}).
\newblock \DOIprefix\doi{https://doi.org/10.1145/3126686.3126742}.
\bibitem[{Ege \& Yanai(2018)}]{DBLP:conf/ijcai/EgeY18}
\bibinfo{author}{Ege, T.}, \& \bibinfo{author}{Yanai, K.}
  (\bibinfo{year}{2018}).
\newblock \bibinfo{title}{Multi-task learning of dish detection and calorie
  estimation}.
\newblock In {\it \bibinfo{booktitle}{Proceedings of the Joint Workshop on
  Multimedia for Cooking and Eating Activities and Multimedia Assisted Dietary
  Management}\/} (pp. \bibinfo{pages}{53--58}).
\newblock \DOIprefix\doi{https://doi.org/10.1145/3230519.3230594}.
\bibitem[{Ege \& Yanai(2019)}]{DBLP:journals/ieicet/EgeY19}
\bibinfo{author}{Ege, T.}, \& \bibinfo{author}{Yanai, K.}
  (\bibinfo{year}{2019}).
\newblock \bibinfo{title}{Simultaneous estimation of dish locations and
  calories with multi-task learning}.
\newblock {\it \bibinfo{journal}{IEICE Transactions on Information and
  Systems}\/},  {\it \bibinfo{volume}{102-D}\/}, \bibinfo{pages}{1240--1246}
  \DOIprefix\doi{https://doi.org/10.1587/transinf.2018CEP0004}.
\bibitem[{Eskin \& Mihailidis(2012)}]{DBLP:conf/aaaifs/EskinM12}
\bibinfo{author}{Eskin, Y.}, \& \bibinfo{author}{Mihailidis, A.}
  (\bibinfo{year}{2012}).
\newblock \bibinfo{title}{An intelligent nutritional assessment system}.
\newblock In {\it \bibinfo{booktitle}{2012 AAAI Fall Symposium Series}\/}.
\newblock \URLprefix
  \url{http://www.aaai.org/ocs/index.php/FSS/FSS12/paper/view/5558}.
\bibitem[{Fang et~al.(2015)Fang, Liu, Zhu, Delp \&
  Boushey}]{DBLP:conf/ism/FangLZDB15}
\bibinfo{author}{Fang, S.}, \bibinfo{author}{Liu, C.}, \bibinfo{author}{Zhu,
  F.}, \bibinfo{author}{Delp, E.~J.}, \& \bibinfo{author}{Boushey, C.~J.}
  (\bibinfo{year}{2015}).
\newblock \bibinfo{title}{Single-view food portion estimation based on
  geometric models}.
\newblock In {\it \bibinfo{booktitle}{2015 {IEEE} International Symposium on
  Multimedia}\/} (pp. \bibinfo{pages}{385--390}).
\newblock \DOIprefix\doi{https://doi.org/10.1109/ISM.2015.67}.
\bibitem[{Fang et~al.(2019)Fang, Shao, Kerr, Boushey \& Zhu}]{fang2019end}
\bibinfo{author}{Fang, S.}, \bibinfo{author}{Shao, Z.}, \bibinfo{author}{Kerr,
  D.~A.}, \bibinfo{author}{Boushey, C.~J.}, \& \bibinfo{author}{Zhu, F.}
  (\bibinfo{year}{2019}).
\newblock \bibinfo{title}{An end-to-end image-based automatic food energy
  estimation technique based on learned energy distribution images: Protocol
  and methodology}.
\newblock {\it \bibinfo{journal}{Nutrients}\/},  {\it \bibinfo{volume}{11}\/}
  \DOIprefix\doi{https://doi.org/10.3390/nu11040877}.
\bibitem[{Fang et~al.(2018)Fang, Shao, Mao, Fu, Delp, Zhu, Kerr \&
  Boushey}]{DBLP:conf/icip/FangSMFDZKB18}
\bibinfo{author}{Fang, S.}, \bibinfo{author}{Shao, Z.}, \bibinfo{author}{Mao,
  R.}, \bibinfo{author}{Fu, C.}, \bibinfo{author}{Delp, E.~J.},
  \bibinfo{author}{Zhu, F.}, \bibinfo{author}{Kerr, D.~A.}, \&
  \bibinfo{author}{Boushey, C.~J.} (\bibinfo{year}{2018}).
\newblock \bibinfo{title}{Single-view food portion estimation: Learning
  image-to-energy mappings using generative adversarial networks}.
\newblock In {\it \bibinfo{booktitle}{2018 {IEEE} International Conference on
  Image Processing}\/} (pp. \bibinfo{pages}{251--255}).
\newblock \DOIprefix\doi{https://doi.org/10.1109/ICIP.2018.8451461}.
\bibitem[{Fontanellaz et~al.(2019)Fontanellaz, Christodoulidis \&
  Mougiakakou}]{10.1145/3347448.3357163}
\bibinfo{author}{Fontanellaz, M.}, \bibinfo{author}{Christodoulidis, S.}, \&
  \bibinfo{author}{Mougiakakou, S.} (\bibinfo{year}{2019}).
\newblock \bibinfo{title}{Self-attention and ingredient-attention based model
  for recipe retrieval from image queries}.
\newblock In {\it \bibinfo{booktitle}{Proceedings of the 5th International
  Workshop on Multimedia Assisted Dietary Management}\/} (p.
  \bibinfo{pages}{25–31}).
\newblock \DOIprefix\doi{https://doi.org/10.1145/3347448.3357163}.
\bibitem[{{Food Databanks National Capability}(2020)}]{FDNC}
\bibinfo{author}{{Food Databanks National Capability}} (\bibinfo{year}{2020}).
\newblock \bibinfo{title}{{Food Databanks National Capability extended dataset
  based on PHE's McCance and Widdowson's Composition of Foods Integrated
  Dataset}}.
\newblock \URLprefix \url{https://quadram.ac.uk/UKfoodcomposition/}
  \bibinfo{note}{. {Accessed} December 9, 2021}.
\bibitem[{Forster et~al.(2014)Forster, Fallaize, Gallagher, O’Donovan,
  Woolhead, Walsh, Macready, Lovegrove, Mathers, Gibney
  et~al.}]{forster2014online}
\bibinfo{author}{Forster, H.}, \bibinfo{author}{Fallaize, R.},
  \bibinfo{author}{Gallagher, C.}, \bibinfo{author}{O’Donovan, C.~B.},
  \bibinfo{author}{Woolhead, C.}, \bibinfo{author}{Walsh, M.~C.},
  \bibinfo{author}{Macready, A.~L.}, \bibinfo{author}{Lovegrove, J.~A.},
  \bibinfo{author}{Mathers, J.~C.}, \bibinfo{author}{Gibney, M.~J.} et~al.
  (\bibinfo{year}{2014}).
\newblock \bibinfo{title}{Online dietary intake estimation: the {Food4Me} food
  frequency questionnaire}.
\newblock {\it \bibinfo{journal}{Journal of Medical Internet Research}\/},
  {\it \bibinfo{volume}{16}\/}, \bibinfo{pages}{e3105}
  \DOIprefix\doi{https://doi.org/10.2196/jmir.3105}.
\bibitem[{Foster et~al.(2008)Foster, O'keeffe, Matthews, Mathers, Nelson,
  Barton, Wrieden \& Adamson}]{foster2008children}
\bibinfo{author}{Foster, E.}, \bibinfo{author}{O'keeffe, M.},
  \bibinfo{author}{Matthews, J.}, \bibinfo{author}{Mathers, J.},
  \bibinfo{author}{Nelson, M.}, \bibinfo{author}{Barton, K.},
  \bibinfo{author}{Wrieden, W.}, \& \bibinfo{author}{Adamson, A.}
  (\bibinfo{year}{2008}).
\newblock \bibinfo{title}{Children's estimates of food portion size: the effect
  of timing of dietary interview on the accuracy of children's portion size
  estimates}.
\newblock {\it \bibinfo{journal}{British Journal of Nutrition}\/},  {\it
  \bibinfo{volume}{99}\/}, \bibinfo{pages}{185--190}
  \DOIprefix\doi{https://doi.org/10.1017/S0007114507791882}.
\bibitem[{Gao et~al.(2018)Gao, Lo \& Lo}]{8329671}
\bibinfo{author}{Gao, A.}, \bibinfo{author}{Lo, F. P.-W.}, \&
  \bibinfo{author}{Lo, B.} (\bibinfo{year}{2018}).
\newblock \bibinfo{title}{Food volume estimation for quantifying dietary intake
  with a wearable camera}.
\newblock In {\it \bibinfo{booktitle}{IEEE 15th International Conference on
  Wearable and Implantable Body Sensor Networks}\/} (pp.
  \bibinfo{pages}{110--113}).
\newblock \DOIprefix\doi{https://doi.org/10.1109/BSN.2018.8329671}.
\bibitem[{Gersovitz et~al.(1978)Gersovitz, Madden \&
  Smiciklas-Wright}]{gersovitz1978validity}
\bibinfo{author}{Gersovitz, M.}, \bibinfo{author}{Madden, J.~P.}, \&
  \bibinfo{author}{Smiciklas-Wright, H.} (\bibinfo{year}{1978}).
\newblock \bibinfo{title}{Validity of the 24-hr. dietary recall and seven-day
  record for group comparisons}.
\newblock {\it \bibinfo{journal}{Journal of the American Dietetic
  Association}\/},  {\it \bibinfo{volume}{73}\/}, \bibinfo{pages}{48--55}
  \URLprefix \url{http://europepmc.org/abstract/MED/659761}.
\bibitem[{Gibney et~al.(2020)Gibney, Allison, Bier \&
  Dwyer}]{gibney2020uncertainty}
\bibinfo{author}{Gibney, M.}, \bibinfo{author}{Allison, D.},
  \bibinfo{author}{Bier, D.}, \& \bibinfo{author}{Dwyer, J.}
  (\bibinfo{year}{2020}).
\newblock \bibinfo{title}{Uncertainty in human nutrition research}.
\newblock {\it \bibinfo{journal}{Nature Food}\/},  {\it \bibinfo{volume}{1}\/},
  \bibinfo{pages}{247--249}
  \DOIprefix\doi{https://doi.org/10.1038/s43016-020-0073-2}.
\bibitem[{{GoBe}(2021)}]{gobe}
\bibinfo{author}{{GoBe}} (\bibinfo{year}{2021}).
\newblock \bibinfo{title}{Healthbe {Gobe} {Automatic} {Body} {Manager}}.
\newblock \URLprefix \url{https://healbe.com/cn/} \bibinfo{note}{. {Accessed}
  August 3, 2021}.
\bibitem[{He et~al.(2021)He, Mao, Shao, Wright, Kerr, Boushey \&
  Zhu}]{DBLP:journals/EndtoEnd}
\bibinfo{author}{He, J.}, \bibinfo{author}{Mao, R.}, \bibinfo{author}{Shao,
  Z.}, \bibinfo{author}{Wright, J.~L.}, \bibinfo{author}{Kerr, D.~A.},
  \bibinfo{author}{Boushey, C.~J.}, \& \bibinfo{author}{Zhu, F.}
  (\bibinfo{year}{2021}).
\newblock \bibinfo{title}{An end-to-end food image analysis system}.
\newblock {\it \bibinfo{journal}{arXiv preprint}\/},  \URLprefix
  \url{https://arxiv.org/abs/2102.00645}.
\bibitem[{He et~al.(2020)He, Shao, Wright, Kerr, Boushey \&
  Zhu}]{DBLP:conf/mipr/HeSWKBZ20}
\bibinfo{author}{He, J.}, \bibinfo{author}{Shao, Z.}, \bibinfo{author}{Wright,
  J.}, \bibinfo{author}{Kerr, D.~A.}, \bibinfo{author}{Boushey, C.~J.}, \&
  \bibinfo{author}{Zhu, F.} (\bibinfo{year}{2020}).
\newblock \bibinfo{title}{Multi-task image-based dietary assessment for food
  recognition and portion size estimation}.
\newblock In {\it \bibinfo{booktitle}{3rd {IEEE} Conference on Multimedia
  Information Processing and Retrieval}\/} (pp. \bibinfo{pages}{49--54}).
\newblock \DOIprefix\doi{https://doi.org/10.1109/MIPR49039.2020.00018}.
\bibitem[{He et~al.(2017)He, Gkioxari, Doll{\'a}r \& Girshick}]{he2017mask}
\bibinfo{author}{He, K.}, \bibinfo{author}{Gkioxari, G.},
  \bibinfo{author}{Doll{\'a}r, P.}, \& \bibinfo{author}{Girshick, R.}
  (\bibinfo{year}{2017}).
\newblock \bibinfo{title}{Mask {R-CNN}}.
\newblock In {\it \bibinfo{booktitle}{{IEEE} International Conference on
  Computer Vision}\/} (pp. \bibinfo{pages}{2961--2969}).
\newblock \DOIprefix\doi{https://doi.org/10.1109/ICCV.2017.322}.
\bibitem[{He et~al.(2016)He, Zhang, Ren \& Sun}]{DBLP:conf/cvpr/HeZRS16}
\bibinfo{author}{He, K.}, \bibinfo{author}{Zhang, X.}, \bibinfo{author}{Ren,
  S.}, \& \bibinfo{author}{Sun, J.} (\bibinfo{year}{2016}).
\newblock \bibinfo{title}{Deep residual learning for image recognition}.
\newblock In {\it \bibinfo{booktitle}{2016 {IEEE} Conference on Computer Vision
  and Pattern Recognition}\/} (pp. \bibinfo{pages}{770--778}).
\newblock \DOIprefix\doi{https://doi.org/10.1109/CVPR.2016.90}.
\bibitem[{He et~al.(2013)He, Xu, Khanna, Boushey \&
  Delp}]{DBLP:conf/icmcs/HeXKBD13}
\bibinfo{author}{He, Y.}, \bibinfo{author}{Xu, C.}, \bibinfo{author}{Khanna,
  N.}, \bibinfo{author}{Boushey, C.~J.}, \& \bibinfo{author}{Delp, E.~J.}
  (\bibinfo{year}{2013}).
\newblock \bibinfo{title}{Food image analysis: Segmentation, identification and
  weight estimation}.
\newblock In {\it \bibinfo{booktitle}{{IEEE} International Conference on
  Multimedia and Expo}\/} (pp. \bibinfo{pages}{1--6}).
\newblock \DOIprefix\doi{https://doi.org/10.1109/ICME.2013.6607548}.
\bibitem[{{Health Canada}(2021)}]{CNF}
\bibinfo{author}{{Health Canada}} (\bibinfo{year}{2021}).
\newblock \bibinfo{title}{{The Canadian Nutrient File}}.
\newblock \URLprefix
  \url{https://www.canada.ca/en/health-canada/services/food-nutrition/healthy-eating/nutrient-data/canadian-nutrient-file-about-us.html}
  \bibinfo{note}{. {Accessed} December 9, 2021}.
\bibitem[{H{\"o}chsmann \& Martin(2020)}]{hochsmann2020review}
\bibinfo{author}{H{\"o}chsmann, C.}, \& \bibinfo{author}{Martin, C.~K.}
  (\bibinfo{year}{2020}).
\newblock \bibinfo{title}{Review of the validity and feasibility of
  image-assisted methods for dietary assessment}.
\newblock {\it \bibinfo{journal}{International Journal of Obesity}\/},  {\it
  \bibinfo{volume}{44}\/}, \bibinfo{pages}{2358--2371}
  \DOIprefix\doi{https://doi.org/10.1038/s41366-020-00693-2}.
\bibitem[{Huang et~al.(2017)Huang, Liu, van~der Maaten \&
  Weinberger}]{DBLP:conf/cvpr/HuangLMW17}
\bibinfo{author}{Huang, G.}, \bibinfo{author}{Liu, Z.},
  \bibinfo{author}{van~der Maaten, L.}, \& \bibinfo{author}{Weinberger, K.~Q.}
  (\bibinfo{year}{2017}).
\newblock \bibinfo{title}{Densely connected convolutional networks}.
\newblock In {\it \bibinfo{booktitle}{Proceedings of the IEEE Conference on
  Computer Vision and Pattern Recognition}\/} (pp.
  \bibinfo{pages}{4700--4708}).
\newblock \DOIprefix\doi{https://doi.org/10.1109/CVPR.2017.243}.
\bibitem[{IARC(2002)}]{international2002iarc}
\bibinfo{author}{IARC} (\bibinfo{year}{2002}).
\newblock \bibinfo{title}{{IARC} handbooks of cancer prevention: Weight control
  and physical activity}.
\newblock {\it \bibinfo{journal}{Lyon, France: International Agency for
  Research on Cancer}\/}, .
\bibitem[{Illner et~al.(2012)Illner, Freisling, Boeing, Huybrechts, Crispim \&
  Slimani}]{illner2012review}
\bibinfo{author}{Illner, A.-K.}, \bibinfo{author}{Freisling, H.},
  \bibinfo{author}{Boeing, H.}, \bibinfo{author}{Huybrechts, I.},
  \bibinfo{author}{Crispim, S.}, \& \bibinfo{author}{Slimani, N.}
  (\bibinfo{year}{2012}).
\newblock \bibinfo{title}{Review and evaluation of innovative technologies for
  measuring diet in nutritional epidemiology}.
\newblock {\it \bibinfo{journal}{International Journal of Epidemiology}\/},
  {\it \bibinfo{volume}{41}\/}, \bibinfo{pages}{1187--1203}
  \DOIprefix\doi{https://doi.org/10.1093/ije/dys105}.
\bibitem[{Ingram et~al.(2020)Ingram, Ajates, Arnall, Blake, Borrelli, Collier,
  de~Frece, H{\"a}sler, Lang, Pope et~al.}]{ingram2020future}
\bibinfo{author}{Ingram, J.}, \bibinfo{author}{Ajates, R.},
  \bibinfo{author}{Arnall, A.}, \bibinfo{author}{Blake, L.},
  \bibinfo{author}{Borrelli, R.}, \bibinfo{author}{Collier, R.},
  \bibinfo{author}{de~Frece, A.}, \bibinfo{author}{H{\"a}sler, B.},
  \bibinfo{author}{Lang, T.}, \bibinfo{author}{Pope, H.} et~al.
  (\bibinfo{year}{2020}).
\newblock \bibinfo{title}{A future workforce of food-system analysts}.
\newblock {\it \bibinfo{journal}{Nature Food}\/},  {\it \bibinfo{volume}{1}\/},
  \bibinfo{pages}{9--10}
  \DOIprefix\doi{https://doi.org/10.1038/s43016-019-0003-3}.
\bibitem[{Jiang et~al.(2020)Jiang, Qiu, Liu, Huang \& Lin}]{jiang2020deepfood}
\bibinfo{author}{Jiang, L.}, \bibinfo{author}{Qiu, B.}, \bibinfo{author}{Liu,
  X.}, \bibinfo{author}{Huang, C.}, \& \bibinfo{author}{Lin, K.}
  (\bibinfo{year}{2020}).
\newblock \bibinfo{title}{Deepfood: food image analysis and dietary assessment
  via deep model}.
\newblock {\it \bibinfo{journal}{IEEE Access}\/},  {\it \bibinfo{volume}{8}\/},
  \bibinfo{pages}{47477--47489}
  \DOIprefix\doi{https://doi.org/10.1109/ACCESS.2020.2973625}.
\bibitem[{Kamilaris \& Prenafeta-Bold{\'u}(2018)}]{kamilaris2018deep}
\bibinfo{author}{Kamilaris, A.}, \& \bibinfo{author}{Prenafeta-Bold{\'u},
  F.~X.} (\bibinfo{year}{2018}).
\newblock \bibinfo{title}{Deep learning in agriculture: A survey}.
\newblock {\it \bibinfo{journal}{Computers and electronics in agriculture}\/},
  {\it \bibinfo{volume}{147}\/}, \bibinfo{pages}{70--90}
  \DOIprefix\doi{https://doi.org/10.1016/j.compag.2018.02.016}.
\bibitem[{Khan et~al.(2021)Khan, Naseer, Hayat, Zamir, Khan \&
  Shah}]{khan2021transformers}
\bibinfo{author}{Khan, S.}, \bibinfo{author}{Naseer, M.},
  \bibinfo{author}{Hayat, M.}, \bibinfo{author}{Zamir, S.~W.},
  \bibinfo{author}{Khan, F.~S.}, \& \bibinfo{author}{Shah, M.}
  (\bibinfo{year}{2021}).
\newblock \bibinfo{title}{Transformers in vision: A survey}.
\newblock {\it \bibinfo{journal}{arXiv preprint}\/},  \URLprefix
  \url{https://arxiv.org/abs/2101.01169}.
\bibitem[{Kirkpatrick et~al.(2014)Kirkpatrick, Subar, Douglass, Zimmerman,
  Thompson, Kahle, George, Dodd \& Potischman}]{kirkpatrick2014performance}
\bibinfo{author}{Kirkpatrick, S.~I.}, \bibinfo{author}{Subar, A.~F.},
  \bibinfo{author}{Douglass, D.}, \bibinfo{author}{Zimmerman, T.~P.},
  \bibinfo{author}{Thompson, F.~E.}, \bibinfo{author}{Kahle, L.~L.},
  \bibinfo{author}{George, S.~M.}, \bibinfo{author}{Dodd, K.~W.}, \&
  \bibinfo{author}{Potischman, N.} (\bibinfo{year}{2014}).
\newblock \bibinfo{title}{Performance of the automated self-administered
  24-hour recall relative to a measure of true intakes and to an
  interviewer-administered 24-h recall}.
\newblock {\it \bibinfo{journal}{The American Journal of Clinical
  Nutrition}\/},  {\it \bibinfo{volume}{100}\/}, \bibinfo{pages}{233--240}
  \DOIprefix\doi{https://doi.org/10.3945/ajcn.114.083238}.
\bibitem[{Kitamura et~al.(2008)Kitamura, Yamasaki \& Aizawa}]{kitamura2008food}
\bibinfo{author}{Kitamura, K.}, \bibinfo{author}{Yamasaki, T.}, \&
  \bibinfo{author}{Aizawa, K.} (\bibinfo{year}{2008}).
\newblock \bibinfo{title}{Food log by analyzing food images}.
\newblock In {\it \bibinfo{booktitle}{Proceedings of the 16th ACM international
  conference on Multimedia}\/} (pp. \bibinfo{pages}{999--1000}).
\newblock \DOIprefix\doi{https://doi.org/10.1145/1459359.1459548}.
\bibitem[{Knez \& {\v{S}}ajn(2020)}]{knez2020food}
\bibinfo{author}{Knez, S.}, \& \bibinfo{author}{{\v{S}}ajn, L.}
  (\bibinfo{year}{2020}).
\newblock \bibinfo{title}{{Food object recognition using a mobile device:
  Evaluation of currently implemented systems}}.
\newblock {\it \bibinfo{journal}{Trends in Food Science \& Technology}\/},
  {\it \bibinfo{volume}{99}\/}, \bibinfo{pages}{460--471}
  \DOIprefix\doi{https://doi.org/10.1016/j.tifs.2020.03.017}.
\bibitem[{Kong \& Tan(2012)}]{DBLP:journals/percom/KongT12}
\bibinfo{author}{Kong, F.}, \& \bibinfo{author}{Tan, J.}
  (\bibinfo{year}{2012}).
\newblock \bibinfo{title}{Dietcam: Automatic dietary assessment with mobile
  camera phones}.
\newblock {\it \bibinfo{journal}{Pervasive and Mobile Computing}\/},  {\it
  \bibinfo{volume}{8}\/}, \bibinfo{pages}{147--163}
  \DOIprefix\doi{https://doi.org/10.1016/j.pmcj.2011.07.003}.
\bibitem[{Kristal et~al.(2014)Kristal, Kolar, Fisher, Plascak, Stumbo, Weiss \&
  Paskett}]{kristal2014evaluation}
\bibinfo{author}{Kristal, A.~R.}, \bibinfo{author}{Kolar, A.~S.},
  \bibinfo{author}{Fisher, J.~L.}, \bibinfo{author}{Plascak, J.~J.},
  \bibinfo{author}{Stumbo, P.~J.}, \bibinfo{author}{Weiss, R.}, \&
  \bibinfo{author}{Paskett, E.~D.} (\bibinfo{year}{2014}).
\newblock \bibinfo{title}{Evaluation of web-based, self-administered, graphical
  food frequency questionnaire}.
\newblock {\it \bibinfo{journal}{Journal of the Academy of Nutrition and
  Dietetics}\/},  {\it \bibinfo{volume}{114}\/}, \bibinfo{pages}{613--621}
  \DOIprefix\doi{https://doi.org/10.1016/j.jand.2013.11.017}.
\bibitem[{Lauby-Secretan et~al.(2016)Lauby-Secretan, Scoccianti, Loomis,
  Grosse, Bianchini \& Straif}]{lauby2016body}
\bibinfo{author}{Lauby-Secretan, B.}, \bibinfo{author}{Scoccianti, C.},
  \bibinfo{author}{Loomis, D.}, \bibinfo{author}{Grosse, Y.},
  \bibinfo{author}{Bianchini, F.}, \& \bibinfo{author}{Straif, K.}
  (\bibinfo{year}{2016}).
\newblock \bibinfo{title}{Body fatness and cancer—viewpoint of the {IARC
  Working Group}}.
\newblock {\it \bibinfo{journal}{New England Journal of Medicine}\/},  {\it
  \bibinfo{volume}{375}\/}, \bibinfo{pages}{794--798}
  \DOIprefix\doi{https://doi.org/10.1056/NEJMsr1606602}.
\bibitem[{LeCun et~al.(2015)LeCun, Bengio \& Hinton}]{lecun2015deep}
\bibinfo{author}{LeCun, Y.}, \bibinfo{author}{Bengio, Y.}, \&
  \bibinfo{author}{Hinton, G.} (\bibinfo{year}{2015}).
\newblock \bibinfo{title}{Deep learning}.
\newblock {\it \bibinfo{journal}{Nature}\/},  {\it \bibinfo{volume}{521}\/},
  \bibinfo{pages}{436--444}
  \DOIprefix\doi{https://doi.org/10.1038/nature14539}.
\bibitem[{Lei et~al.(2021)Lei, Qiu, Lo \& Lo}]{lei2021assessing}
\bibinfo{author}{Lei, J.}, \bibinfo{author}{Qiu, J.}, \bibinfo{author}{Lo, F.
  P.-W.}, \& \bibinfo{author}{Lo, B.} (\bibinfo{year}{2021}).
\newblock \bibinfo{title}{Assessing individual dietary intake in food sharing
  scenarios with food and human pose detection}.
\newblock In {\it \bibinfo{booktitle}{International Conference on Pattern
  Recognition}\/} (pp. \bibinfo{pages}{549--557}).
\newblock \DOIprefix\doi{https://doi.org/10.1007/978-3-030-68821-9_45}.
\bibitem[{Liu et~al.(2018)Liu, Cao, Luo, Chen, Vokkarane, Yunsheng, Chen \&
  Hou}]{DBLP:journals/tsc/LiuCLCVMCH18}
\bibinfo{author}{Liu, C.}, \bibinfo{author}{Cao, Y.}, \bibinfo{author}{Luo,
  Y.}, \bibinfo{author}{Chen, G.}, \bibinfo{author}{Vokkarane, V.},
  \bibinfo{author}{Yunsheng, M.}, \bibinfo{author}{Chen, S.}, \&
  \bibinfo{author}{Hou, P.} (\bibinfo{year}{2018}).
\newblock \bibinfo{title}{A new deep learning-based food recognition system for
  dietary assessment on an edge computing service infrastructure}.
\newblock {\it \bibinfo{journal}{IEEE Transactions on Services Computing}\/},
  {\it \bibinfo{volume}{11}\/}, \bibinfo{pages}{249--261}
  \DOIprefix\doi{https://doi.org/10.1109/TSC.2017.2662008}.
\bibitem[{Liu et~al.(2021)Liu, Pu \& Sun}]{liu2021efficient}
\bibinfo{author}{Liu, Y.}, \bibinfo{author}{Pu, H.}, \& \bibinfo{author}{Sun,
  D.-W.} (\bibinfo{year}{2021}).
\newblock \bibinfo{title}{Efficient extraction of deep image features using
  convolutional neural network ({CNN}) for applications in detecting and
  analysing complex food matrices}.
\newblock {\it \bibinfo{journal}{Trends in Food Science \& Technology}\/},
  {\it \bibinfo{volume}{113}\/}, \bibinfo{pages}{193--204}
  \DOIprefix\doi{https://doi.org/10.1016/j.tifs.2021.04.042}.
\bibitem[{Lo et~al.(2020{\natexlab{a}})Lo, Sun, Qiu \&
  Lo}]{DBLP:journals/titb/LoSQL20}
\bibinfo{author}{Lo, F. P.~W.}, \bibinfo{author}{Sun, Y.},
  \bibinfo{author}{Qiu, J.}, \& \bibinfo{author}{Lo, B.}
  (\bibinfo{year}{2020}{\natexlab{a}}).
\newblock \bibinfo{title}{Image-based food classification and volume estimation
  for dietary assessment: {A} review}.
\newblock {\it \bibinfo{journal}{IEEE Journal of Biomedical and Health
  Informatics}\/},  {\it \bibinfo{volume}{24}\/}, \bibinfo{pages}{1926--1939}
  \DOIprefix\doi{https://doi.org/10.1109/JBHI.2020.2987943}.
\bibitem[{Lo et~al.(2020{\natexlab{b}})Lo, Sun, Qiu \&
  Lo}]{DBLP:journals/tii/LoSQL20}
\bibinfo{author}{Lo, F. P.~W.}, \bibinfo{author}{Sun, Y.},
  \bibinfo{author}{Qiu, J.}, \& \bibinfo{author}{Lo, B. P.~L.}
  (\bibinfo{year}{2020}{\natexlab{b}}).
\newblock \bibinfo{title}{Point2volume: {A} vision-based dietary assessment
  approach using view synthesis}.
\newblock {\it \bibinfo{journal}{IEEE Transactions on Industrial
  Informatics}\/},  {\it \bibinfo{volume}{16}\/}, \bibinfo{pages}{577--586}
  \DOIprefix\doi{https://doi.org/10.1109/TII.2019.2942831}.
\bibitem[{Lu et~al.(2018)Lu, Allegra, Anthimopoulos, Stanco, Farinella \&
  Mougiakakou}]{DBLP:conf/ijcai/LuAASFM18}
\bibinfo{author}{Lu, Y.}, \bibinfo{author}{Allegra, D.},
  \bibinfo{author}{Anthimopoulos, M.}, \bibinfo{author}{Stanco, F.},
  \bibinfo{author}{Farinella, G.~M.}, \& \bibinfo{author}{Mougiakakou, S.~G.}
  (\bibinfo{year}{2018}).
\newblock \bibinfo{title}{A multi-task learning approach for meal assessment}.
\newblock In {\it \bibinfo{booktitle}{Proceedings of the Joint Workshop on
  Multimedia for Cooking and Eating Activities and Multimedia Assisted Dietary
  Management}\/} (pp. \bibinfo{pages}{46--52}).
\newblock \DOIprefix\doi{https://doi.org/10.1145/3230519.3230593}.
\bibitem[{Lu et~al.(2021{\natexlab{a}})Lu, Stathopoulou \&
  Mougiakakou}]{DBLP:journals/corr/abs-2008-00818}
\bibinfo{author}{Lu, Y.}, \bibinfo{author}{Stathopoulou, T.}, \&
  \bibinfo{author}{Mougiakakou, S.} (\bibinfo{year}{2021}{\natexlab{a}}).
\newblock \bibinfo{title}{Partially supervised multi-task network for
  single-view dietary assessment}.
\newblock In {\it \bibinfo{booktitle}{International Conference on Pattern
  Recognition}\/} (pp. \bibinfo{pages}{8156--8163}).
\newblock \DOIprefix\doi{https://doi.org/10.1109/ICPR48806.2021.9412339}.
\bibitem[{Lu et~al.(2021{\natexlab{b}})Lu, Stathopoulou, Vasiloglou,
  Christodoulidis, Stanga \& Mougiakakou}]{DBLP:journals/corr/abs-2003-08273}
\bibinfo{author}{Lu, Y.}, \bibinfo{author}{Stathopoulou, T.},
  \bibinfo{author}{Vasiloglou, M.~F.}, \bibinfo{author}{Christodoulidis, S.},
  \bibinfo{author}{Stanga, Z.}, \& \bibinfo{author}{Mougiakakou, S.}
  (\bibinfo{year}{2021}{\natexlab{b}}).
\newblock \bibinfo{title}{An artificial intelligence-based system to assess
  nutrient intake for hospitalised patients}.
\newblock {\it \bibinfo{journal}{IEEE Transactions on Multimedia}\/},  {\it
  \bibinfo{volume}{23}\/}, \bibinfo{pages}{1136--1147}
  \DOIprefix\doi{https://doi.org/10.1109/TMM.2020.2993948}.
\bibitem[{Lu et~al.(2020)Lu, Stathopoulou, Vasiloglou, Pinault, Kiley, Spanakis
  \& Mougiakakou}]{goFOODTM}
\bibinfo{author}{Lu, Y.}, \bibinfo{author}{Stathopoulou, T.},
  \bibinfo{author}{Vasiloglou, M.~F.}, \bibinfo{author}{Pinault, L.~F.},
  \bibinfo{author}{Kiley, C.}, \bibinfo{author}{Spanakis, E.~K.}, \&
  \bibinfo{author}{Mougiakakou, S.} (\bibinfo{year}{2020}).
\newblock \bibinfo{title}{{goFOODTM}: An artificial intelligence system for
  dietary assessment}.
\newblock {\it \bibinfo{journal}{Sensors}\/},  {\it \bibinfo{volume}{20}\/},
  \bibinfo{pages}{4283} \DOIprefix\doi{https://doi.org/10.3390/s20154283}.
\bibitem[{Ma et~al.(2021)Ma, Lau, Yu, Li, Liu, Wang \& Sheng}]{ma2021image}
\bibinfo{author}{Ma, P.}, \bibinfo{author}{Lau, C.~P.}, \bibinfo{author}{Yu,
  N.}, \bibinfo{author}{Li, A.}, \bibinfo{author}{Liu, P.},
  \bibinfo{author}{Wang, Q.}, \& \bibinfo{author}{Sheng, J.}
  (\bibinfo{year}{2021}).
\newblock \bibinfo{title}{Image-based nutrient estimation for {Chinese} dishes
  using deep learning}.
\newblock {\it \bibinfo{journal}{Food Research International}\/},  {\it
  \bibinfo{volume}{147}\/}, \bibinfo{pages}{110437}
  \DOIprefix\doi{https://doi.org/10.1016/j.foodres.2021.110437}.
\bibitem[{Ma et~al.(2022)Ma, Lau, Yu, Li \& Sheng}]{ma2022application}
\bibinfo{author}{Ma, P.}, \bibinfo{author}{Lau, C.~P.}, \bibinfo{author}{Yu,
  N.}, \bibinfo{author}{Li, A.}, \& \bibinfo{author}{Sheng, J.}
  (\bibinfo{year}{2022}).
\newblock \bibinfo{title}{Application of deep learning for image-based chinese
  market food nutrients estimation}.
\newblock {\it \bibinfo{journal}{Food Chemistry}\/},  {\it
  \bibinfo{volume}{373}\/}, \bibinfo{pages}{130994}
  \DOIprefix\doi{https://doi.org/10.1016/j.foodchem.2021.130994}.
\bibitem[{Mao et~al.(2021)Mao, He, Shao, Yarlagadda \& Zhu}]{mao2021visual}
\bibinfo{author}{Mao, R.}, \bibinfo{author}{He, J.}, \bibinfo{author}{Shao,
  Z.}, \bibinfo{author}{Yarlagadda, S.~K.}, \& \bibinfo{author}{Zhu, F.}
  (\bibinfo{year}{2021}).
\newblock \bibinfo{title}{Visual aware hierarchy based food recognition}.
\newblock In {\it \bibinfo{booktitle}{International Conference on Pattern
  Recognition}\/} (pp. \bibinfo{pages}{571--598}).
\newblock \DOIprefix\doi{https://doi.org/10.1007/978-3-030-68821-9\_47}.
\bibitem[{Mariappan et~al.(2009)Mariappan, Bosch, Zhu, Boushey, Kerr, Ebert \&
  Delp}]{DBLP:conf/cimaging/MariappanBZBKED09}
\bibinfo{author}{Mariappan, A.}, \bibinfo{author}{Bosch, M.},
  \bibinfo{author}{Zhu, F.}, \bibinfo{author}{Boushey, C.~J.},
  \bibinfo{author}{Kerr, D.~A.}, \bibinfo{author}{Ebert, D.~S.}, \&
  \bibinfo{author}{Delp, E.~J.} (\bibinfo{year}{2009}).
\newblock \bibinfo{title}{Personal dietary assessment using mobile devices}.
\newblock In {\it \bibinfo{booktitle}{Computational Imaging VII}\/} (p.
  \bibinfo{pages}{72460Z}).
\newblock \DOIprefix\doi{https://doi.org/10.1117/12.813556}.
\bibitem[{McDonald et~al.(2016)McDonald, Glusman \&
  Price}]{mcdonald2016personalized}
\bibinfo{author}{McDonald, D.}, \bibinfo{author}{Glusman, G.}, \&
  \bibinfo{author}{Price, N.~D.} (\bibinfo{year}{2016}).
\newblock \bibinfo{title}{Personalized nutrition through big data}.
\newblock {\it \bibinfo{journal}{Nature biotechnology}\/},  {\it
  \bibinfo{volume}{34}\/}, \bibinfo{pages}{152--154}.
\bibitem[{McPherson et~al.(2000)McPherson, Hoelscher, Alexander, Scanlon \&
  Serdula}]{mcpherson2000dietary}
\bibinfo{author}{McPherson, R.~S.}, \bibinfo{author}{Hoelscher, D.~M.},
  \bibinfo{author}{Alexander, M.}, \bibinfo{author}{Scanlon, K.~S.}, \&
  \bibinfo{author}{Serdula, M.~K.} (\bibinfo{year}{2000}).
\newblock \bibinfo{title}{Dietary assessment methods among school-aged
  children: validity and reliability}.
\newblock {\it \bibinfo{journal}{Preventive Medicine}\/},  {\it
  \bibinfo{volume}{31}\/}, \bibinfo{pages}{S11--S33}
  \DOIprefix\doi{https://doi.org/10.1006/pmed.2000.0631}.
\bibitem[{Mezgec \& Korou{\v{s}}i{\'c}~Seljak(2017)}]{mezgec2017nutrinet}
\bibinfo{author}{Mezgec, S.}, \& \bibinfo{author}{Korou{\v{s}}i{\'c}~Seljak,
  B.} (\bibinfo{year}{2017}).
\newblock \bibinfo{title}{Nutrinet: a deep learning food and drink image
  recognition system for dietary assessment}.
\newblock {\it \bibinfo{journal}{Nutrients}\/},  {\it \bibinfo{volume}{9}\/},
  \bibinfo{pages}{657} \DOIprefix\doi{https://doi.org/10.3390/nu9070657}.
\bibitem[{Min et~al.(2020{\natexlab{a}})Min, Jiang \& Jain}]{8930090}
\bibinfo{author}{Min, W.}, \bibinfo{author}{Jiang, S.}, \&
  \bibinfo{author}{Jain, R.} (\bibinfo{year}{2020}{\natexlab{a}}).
\newblock \bibinfo{title}{Food recommendation: Framework, existing solutions,
  and challenges}.
\newblock {\it \bibinfo{journal}{IEEE Transactions on Multimedia}\/},  {\it
  \bibinfo{volume}{22}\/}, \bibinfo{pages}{2659--2671}
  \DOIprefix\doi{https://doi.org/10.1109/TMM.2019.2958761}.
\bibitem[{Min et~al.(2019)Min, Jiang, Liu, Rui \&
  Jain}]{DBLP:journals/csur/MinJLRJ19}
\bibinfo{author}{Min, W.}, \bibinfo{author}{Jiang, S.}, \bibinfo{author}{Liu,
  L.}, \bibinfo{author}{Rui, Y.}, \& \bibinfo{author}{Jain, R.~C.}
  (\bibinfo{year}{2019}).
\newblock \bibinfo{title}{A survey on food computing}.
\newblock {\it \bibinfo{journal}{ACM Computing Surveys}\/},  {\it
  \bibinfo{volume}{52}\/}, \bibinfo{pages}{92:1--92:36}
  \DOIprefix\doi{https://doi.org/10.1145/3329168}.
\bibitem[{Min et~al.(2020{\natexlab{b}})Min, Liu, Wang, Luo, Wei, Wei \&
  Jiang}]{Weiqing-Food500-MM2020}
\bibinfo{author}{Min, W.}, \bibinfo{author}{Liu, L.}, \bibinfo{author}{Wang,
  Z.}, \bibinfo{author}{Luo, Z.}, \bibinfo{author}{Wei, X.},
  \bibinfo{author}{Wei, X.}, \& \bibinfo{author}{Jiang, S.}
  (\bibinfo{year}{2020}{\natexlab{b}}).
\newblock \bibinfo{title}{{ISIA} food-500: {A} dataset for large-scale food
  recognition via stacked global-local attention network}.
\newblock In {\it \bibinfo{booktitle}{{ACM} Multimedia}\/} (pp.
  \bibinfo{pages}{393--401}).
\newblock \DOIprefix\doi{https://doi.org/10.1145/3394171.3414031}.
\bibitem[{Min et~al.(2021)Min, Wang, Liu, Luo, Kang, Wei, Wei \&
  Jiang}]{Weiqing-LSVFR-CoRR2021}
\bibinfo{author}{Min, W.}, \bibinfo{author}{Wang, Z.}, \bibinfo{author}{Liu,
  Y.}, \bibinfo{author}{Luo, M.}, \bibinfo{author}{Kang, L.},
  \bibinfo{author}{Wei, X.}, \bibinfo{author}{Wei, X.}, \&
  \bibinfo{author}{Jiang, S.} (\bibinfo{year}{2021}).
\newblock \bibinfo{title}{Large scale visual food recognition}.
\newblock {\it \bibinfo{journal}{arXiv preprint}\/},  \URLprefix
  \url{https://arxiv.org/abs/2103.16107}.
\bibitem[{Minaee et~al.(2021)Minaee, Boykov, Porikli, Plaza, Kehtarnavaz \&
  Terzopoulos}]{minaee2021image}
\bibinfo{author}{Minaee, S.}, \bibinfo{author}{Boykov, Y.~Y.},
  \bibinfo{author}{Porikli, F.}, \bibinfo{author}{Plaza, A.~J.},
  \bibinfo{author}{Kehtarnavaz, N.}, \& \bibinfo{author}{Terzopoulos, D.}
  (\bibinfo{year}{2021}).
\newblock \bibinfo{title}{Image segmentation using deep learning: A survey}.
\newblock {\it \bibinfo{journal}{IEEE Transactions on Pattern Analysis and
  Machine Intelligence}\/},  (pp. \bibinfo{pages}{1--1})
  \DOIprefix\doi{https://doi.org/10.1109/TPAMI.2021.3059968}.
\bibitem[{Miyazaki et~al.(2011)Miyazaki, de~Silva \&
  Aizawa}]{miyazaki2011image}
\bibinfo{author}{Miyazaki, T.}, \bibinfo{author}{de~Silva, G.~C.}, \&
  \bibinfo{author}{Aizawa, K.} (\bibinfo{year}{2011}).
\newblock \bibinfo{title}{Image-based calorie content estimation for dietary
  assessment}.
\newblock In {\it \bibinfo{booktitle}{{IEEE} International Symposium on
  Multimedia}\/} (pp. \bibinfo{pages}{363--368}).
\newblock \DOIprefix\doi{https://doi.org/10.1109/ISM.2011.66}.
\bibitem[{Myers et~al.(2015)Myers, Johnston, Rathod, Korattikara, Gorban,
  Silberman, Guadarrama, Papandreou, Huang \&
  Murphy}]{DBLP:conf/iccv/MyersJRKGSGPH015}
\bibinfo{author}{Myers, A.}, \bibinfo{author}{Johnston, N.},
  \bibinfo{author}{Rathod, V.}, \bibinfo{author}{Korattikara, A.},
  \bibinfo{author}{Gorban, A.~N.}, \bibinfo{author}{Silberman, N.},
  \bibinfo{author}{Guadarrama, S.}, \bibinfo{author}{Papandreou, G.},
  \bibinfo{author}{Huang, J.}, \& \bibinfo{author}{Murphy, K.}
  (\bibinfo{year}{2015}).
\newblock \bibinfo{title}{Im2calories: Towards an automated mobile vision food
  diary}.
\newblock In {\it \bibinfo{booktitle}{{IEEE} International Conference on
  Computer Vision}\/} (pp. \bibinfo{pages}{1233--1241}).
\newblock \DOIprefix\doi{https://doi.org/10.1109/ICCV.2015.146}.
\bibitem[{Nordstr{\"o}m et~al.(2013)Nordstr{\"o}m, Coff, J{\"o}nsson,
  Nordenfelt \& G{\"o}rman}]{nordstrom2013food}
\bibinfo{author}{Nordstr{\"o}m, K.}, \bibinfo{author}{Coff, C.},
  \bibinfo{author}{J{\"o}nsson, H.}, \bibinfo{author}{Nordenfelt, L.}, \&
  \bibinfo{author}{G{\"o}rman, U.} (\bibinfo{year}{2013}).
\newblock \bibinfo{title}{Food and health: individual, cultural, or scientific
  matters?}
\newblock {\it \bibinfo{journal}{Genes \& nutrition}\/},  {\it
  \bibinfo{volume}{8}\/}, \bibinfo{pages}{357--363}
  \DOIprefix\doi{https://doi.org/10.1007/s12263-013-0336-8}.
\bibitem[{Okamoto \& Yanai(2016{\natexlab{a}})}]{DBLP:conf/mm/OkamotoY16}
\bibinfo{author}{Okamoto, K.}, \& \bibinfo{author}{Yanai, K.}
  (\bibinfo{year}{2016}{\natexlab{a}}).
\newblock \bibinfo{title}{An automatic calorie estimation system of food images
  on a smartphone}.
\newblock In {\it \bibinfo{booktitle}{Proceedings of the 2nd International
  Workshop on Multimedia Assisted Dietary Management}\/} (pp.
  \bibinfo{pages}{63--70}).
\newblock \DOIprefix\doi{https://doi.org/10.1145/2986035.2986040}.
\bibitem[{Okamoto \& Yanai(2016{\natexlab{b}})}]{DBLP:conf/mmm/OkamotoY16}
\bibinfo{author}{Okamoto, K.}, \& \bibinfo{author}{Yanai, K.}
  (\bibinfo{year}{2016}{\natexlab{b}}).
\newblock \bibinfo{title}{Grillcam: {A} real-time eating action recognition
  system}.
\newblock In {\it \bibinfo{booktitle}{International Conference on Multimedia
  Modeling}\/} (pp. \bibinfo{pages}{331--335}).
\newblock \DOIprefix\doi{https://doi.org/10.1007/978-3-319-27674-8\_29}.
\bibitem[{{Open Food Facts}(2020)}]{OFF}
\bibinfo{author}{{Open Food Facts}} (\bibinfo{year}{2020}).
\newblock \bibinfo{title}{{Open Food Facts: the free food products database}}.
\newblock \URLprefix \url{https://world.openfoodfacts.org/} \bibinfo{note}{.
  {Accessed} December 9, 2021}.
\bibitem[{Pouladzadeh et~al.(2016{\natexlab{a}})Pouladzadeh, Kuhad, Peddi,
  Yassine \& Shirmohammadi}]{DBLP:conf/i2mtc/PouladzadehKPYS16}
\bibinfo{author}{Pouladzadeh, P.}, \bibinfo{author}{Kuhad, P.},
  \bibinfo{author}{Peddi, S. V.~B.}, \bibinfo{author}{Yassine, A.}, \&
  \bibinfo{author}{Shirmohammadi, S.} (\bibinfo{year}{2016}{\natexlab{a}}).
\newblock \bibinfo{title}{Food calorie measurement using deep learning neural
  network}.
\newblock In {\it \bibinfo{booktitle}{{IEEE} International Instrumentation and
  Measurement Technology Conference}\/} (pp. \bibinfo{pages}{1--6}).
\newblock \DOIprefix\doi{https://doi.org/10.1109/I2MTC.2016.7520547}.
\bibitem[{Pouladzadeh \&
  Shirmohammadi(2017)}]{DBLP:journals/tomccap/PouladzadehS17}
\bibinfo{author}{Pouladzadeh, P.}, \& \bibinfo{author}{Shirmohammadi, S.}
  (\bibinfo{year}{2017}).
\newblock \bibinfo{title}{Mobile multi-food recognition using deep learning}.
\newblock {\it \bibinfo{journal}{ACM Transactions on Multimedia Computing
  Communications and Applications}\/},  {\it \bibinfo{volume}{13}\/},
  \bibinfo{pages}{36:1--36:21} \DOIprefix\doi{https://doi.org/10.1145/3063592}.
\bibitem[{Pouladzadeh et~al.(2014{\natexlab{a}})Pouladzadeh, Shirmohammadi \&
  Almaghrabi}]{DBLP:journals/tim/PouladzadehSA14}
\bibinfo{author}{Pouladzadeh, P.}, \bibinfo{author}{Shirmohammadi, S.}, \&
  \bibinfo{author}{Almaghrabi, R.} (\bibinfo{year}{2014}{\natexlab{a}}).
\newblock \bibinfo{title}{Measuring calorie and nutrition from food image}.
\newblock {\it \bibinfo{journal}{IEEE Transactions on Instrumentation and
  Measurement}\/},  {\it \bibinfo{volume}{63}\/}, \bibinfo{pages}{1947--1956}
  \DOIprefix\doi{https://doi.org/10.1109/TIM.2014.2303533}.
\bibitem[{Pouladzadeh et~al.(2014{\natexlab{b}})Pouladzadeh, Shirmohammadi \&
  Yassine}]{DBLP:conf/memea/PouladzadehSY14}
\bibinfo{author}{Pouladzadeh, P.}, \bibinfo{author}{Shirmohammadi, S.}, \&
  \bibinfo{author}{Yassine, A.} (\bibinfo{year}{2014}{\natexlab{b}}).
\newblock \bibinfo{title}{Using graph cut segmentation for food calorie
  measurement}.
\newblock In {\it \bibinfo{booktitle}{{IEEE} International Symposium on Medical
  Measurements and Applications}\/} (pp. \bibinfo{pages}{621--626}).
\newblock \DOIprefix\doi{https://doi.org/10.1109/MeMeA.2014.6860137}.
\bibitem[{Pouladzadeh et~al.(2016{\natexlab{b}})Pouladzadeh, Shirmohammadi \&
  Yassine}]{DBLP:journals/imm/PouladzadehSY16}
\bibinfo{author}{Pouladzadeh, P.}, \bibinfo{author}{Shirmohammadi, S.}, \&
  \bibinfo{author}{Yassine, A.} (\bibinfo{year}{2016}{\natexlab{b}}).
\newblock \bibinfo{title}{You are what you eat: So measure what you eat!}
\newblock {\it \bibinfo{journal}{IEEE Instrumentation \& Measurement
  Magazine}\/},  {\it \bibinfo{volume}{19}\/}, \bibinfo{pages}{9--15}
  \DOIprefix\doi{https://doi.org/10.1109/MIM.2016.7384954}.
\bibitem[{Pouladzadeh et~al.(2015)Pouladzadeh, Yassine \&
  Shirmohammadi}]{DBLP:conf/iciap/PouladzadehYS15}
\bibinfo{author}{Pouladzadeh, P.}, \bibinfo{author}{Yassine, A.}, \&
  \bibinfo{author}{Shirmohammadi, S.} (\bibinfo{year}{2015}).
\newblock \bibinfo{title}{{FooDD}: Food detection dataset for calorie
  measurement using food images}.
\newblock In {\it \bibinfo{booktitle}{International Conference on Image
  Analysis and Processing}\/} (pp. \bibinfo{pages}{441--448}).
\newblock \DOIprefix\doi{https://doi.org/10.1007/978-3-319-23222-5\_54}.
\bibitem[{Probst et~al.(2015)Probst, Nguyen, Tran \& Li}]{probst2015dietary}
\bibinfo{author}{Probst, Y.}, \bibinfo{author}{Nguyen, D.~T.},
  \bibinfo{author}{Tran, M.~K.}, \& \bibinfo{author}{Li, W.}
  (\bibinfo{year}{2015}).
\newblock \bibinfo{title}{Dietary assessment on a mobile phone using image
  processing and pattern recognition techniques: algorithm design and system
  prototyping}.
\newblock {\it \bibinfo{journal}{Nutrients}\/},  {\it \bibinfo{volume}{7}\/},
  \bibinfo{pages}{6128--6138}
  \DOIprefix\doi{https://doi.org/10.3390/nu7085274}.
\bibitem[{Puri et~al.(2009)Puri, Zhu, Yu, Divakaran \&
  Sawhney}]{DBLP:conf/wacv/PuriZYDS09}
\bibinfo{author}{Puri, M.}, \bibinfo{author}{Zhu, Z.}, \bibinfo{author}{Yu,
  Q.}, \bibinfo{author}{Divakaran, A.}, \& \bibinfo{author}{Sawhney, H.}
  (\bibinfo{year}{2009}).
\newblock \bibinfo{title}{Recognition and volume estimation of food intake
  using a mobile device}.
\newblock In {\it \bibinfo{booktitle}{2009 Workshop on Applications of Computer
  Vision}\/} (pp. \bibinfo{pages}{1--8}).
\newblock \DOIprefix\doi{https://doi.org/10.1109/WACV.2009.5403087}.
\bibitem[{Qiu et~al.(2019)Qiu, Lo \& Lo}]{DBLP:conf/bsn/QiuLL19}
\bibinfo{author}{Qiu, J.}, \bibinfo{author}{Lo, F. P.-W.}, \&
  \bibinfo{author}{Lo, B.} (\bibinfo{year}{2019}).
\newblock \bibinfo{title}{Assessing individual dietary intake in food sharing
  scenarios with a 360 camera and deep learning}.
\newblock In {\it \bibinfo{booktitle}{IEEE International Conference on Wearable
  and Implantable Body Sensor Networks}\/} (pp. \bibinfo{pages}{1--4}).
\newblock \DOIprefix\doi{https://doi.org/10.1109/BSN.2019.8771095}.
\bibitem[{Rahman et~al.(2012)Rahman, Li, Pickering, Frater, Kerr, Boushey \&
  Delp}]{DBLP:conf/sitis/RahmanLPFKBD12}
\bibinfo{author}{Rahman, M.~H.}, \bibinfo{author}{Li, Q.},
  \bibinfo{author}{Pickering, M.~R.}, \bibinfo{author}{Frater, M.~R.},
  \bibinfo{author}{Kerr, D.~A.}, \bibinfo{author}{Boushey, C.~J.}, \&
  \bibinfo{author}{Delp, E.~J.} (\bibinfo{year}{2012}).
\newblock \bibinfo{title}{Food volume estimation in a mobile phone based
  dietary assessment system}.
\newblock In {\it \bibinfo{booktitle}{Eighth International Conference on Signal
  Image Technology and Internet Based Systems}\/} (pp.
  \bibinfo{pages}{988--995}).
\newblock \DOIprefix\doi{https://doi.org/10.1109/SITIS.2012.146}.
\bibitem[{Redmon et~al.(2016)Redmon, Divvala, Girshick \&
  Farhadi}]{DBLP:conf/cvpr/RedmonDGF16}
\bibinfo{author}{Redmon, J.}, \bibinfo{author}{Divvala, S.~K.},
  \bibinfo{author}{Girshick, R.~B.}, \& \bibinfo{author}{Farhadi, A.}
  (\bibinfo{year}{2016}).
\newblock \bibinfo{title}{You only look once: Unified, real-time object
  detection}.
\newblock In {\it \bibinfo{booktitle}{{IEEE} Conference on Computer Vision and
  Pattern Recognition}\/} (pp. \bibinfo{pages}{779--788}).
\newblock \DOIprefix\doi{https://doi.org/10.1109/CVPR.2016.91}.
\bibitem[{Ren et~al.(2017)Ren, He, Girshick \& Sun}]{7485869}
\bibinfo{author}{Ren, S.}, \bibinfo{author}{He, K.}, \bibinfo{author}{Girshick,
  R.}, \& \bibinfo{author}{Sun, J.} (\bibinfo{year}{2017}).
\newblock \bibinfo{title}{{Faster R-CNN}: Towards real-time object detection
  with region proposal networks}.
\newblock {\it \bibinfo{journal}{IEEE Transactions on Pattern Analysis and
  Machine Intelligence}\/},  {\it \bibinfo{volume}{39}\/},
  \bibinfo{pages}{1137--1149}
  \DOIprefix\doi{https://doi.org/10.1109/TPAMI.2016.2577031}.
\bibitem[{Ronneberger et~al.(2015)Ronneberger, Fischer \&
  Brox}]{ronneberger2015u}
\bibinfo{author}{Ronneberger, O.}, \bibinfo{author}{Fischer, P.}, \&
  \bibinfo{author}{Brox, T.} (\bibinfo{year}{2015}).
\newblock \bibinfo{title}{{U-Net}: Convolutional networks for biomedical image
  segmentation}.
\newblock In {\it \bibinfo{booktitle}{International Conference on Medical image
  computing and computer-assisted intervention}\/} (pp.
  \bibinfo{pages}{234--241}).
\bibitem[{Ruder(2017)}]{ruder2017overview}
\bibinfo{author}{Ruder, S.} (\bibinfo{year}{2017}).
\newblock \bibinfo{title}{An overview of multi-task learning in deep neural
  networks}.
\newblock {\it \bibinfo{journal}{arXiv preprint}\/},  \URLprefix
  \url{https://arxiv.org/abs/1706.05098}.
\bibitem[{Ruede et~al.(2021)Ruede, Heusser, Frank, Roitberg, Haurilet \&
  Stiefelhagen}]{DBLP:journals/corr/abs-2011-01082}
\bibinfo{author}{Ruede, R.}, \bibinfo{author}{Heusser, V.},
  \bibinfo{author}{Frank, L.}, \bibinfo{author}{Roitberg, A.},
  \bibinfo{author}{Haurilet, M.}, \& \bibinfo{author}{Stiefelhagen, R.}
  (\bibinfo{year}{2021}).
\newblock \bibinfo{title}{Multi-task learning for calorie prediction on a novel
  large-scale recipe dataset enriched with nutritional information}.
\newblock In {\it \bibinfo{booktitle}{International Conference on Pattern
  Recognition}\/} (pp. \bibinfo{pages}{4001--4008}).
\newblock \DOIprefix\doi{https://doi.org/10.1109/ICPR48806.2021.9412839}.
\bibitem[{Shao et~al.(2021)Shao, Fang, Mao, He, Wright, Kerr, Boushey \&
  Zhu}]{DBLP:journals/corr/abs-2103-07562}
\bibinfo{author}{Shao, Z.}, \bibinfo{author}{Fang, S.}, \bibinfo{author}{Mao,
  R.}, \bibinfo{author}{He, J.}, \bibinfo{author}{Wright, J.},
  \bibinfo{author}{Kerr, D.}, \bibinfo{author}{Boushey, C.~J.}, \&
  \bibinfo{author}{Zhu, F.} (\bibinfo{year}{2021}).
\newblock \bibinfo{title}{Towards learning food portion from monocular images
  with cross-domain feature adaptation}.
\newblock {\it \bibinfo{journal}{arXiv preprint}\/},  \URLprefix
  \url{https://arxiv.org/abs/2103.07562}.
\bibitem[{Shim et~al.(2014)Shim, Oh \& Kim}]{shim2014dietary}
\bibinfo{author}{Shim, J.-S.}, \bibinfo{author}{Oh, K.}, \&
  \bibinfo{author}{Kim, H.~C.} (\bibinfo{year}{2014}).
\newblock \bibinfo{title}{Dietary assessment methods in epidemiologic studies}.
\newblock {\it \bibinfo{journal}{Epidemiology and Health}\/},  {\it
  \bibinfo{volume}{36}\/}
  \DOIprefix\doi{https://doi.org/10.4178/epih/e2014009}.
\bibitem[{Shroff et~al.(2008)Shroff, Smailagic \&
  Siewiorek}]{DBLP:conf/iswc/ShroffSS08}
\bibinfo{author}{Shroff, G.}, \bibinfo{author}{Smailagic, A.}, \&
  \bibinfo{author}{Siewiorek, D.~P.} (\bibinfo{year}{2008}).
\newblock \bibinfo{title}{Wearable context-aware food recognition for calorie
  monitoring}.
\newblock In {\it \bibinfo{booktitle}{{IEEE} International Symposium on
  Wearable Computers}\/} (pp. \bibinfo{pages}{119--120}).
\newblock \DOIprefix\doi{https://doi.org/10.1109/ISWC.2008.4911602}.
\bibitem[{Situju et~al.(2019)Situju, Takimoto, Sato, Yamauchi, Kanagawa \&
  Lawi}]{DBLP:journals/aai/SitujuTSYKL19}
\bibinfo{author}{Situju, S.~F.}, \bibinfo{author}{Takimoto, H.},
  \bibinfo{author}{Sato, S.}, \bibinfo{author}{Yamauchi, H.},
  \bibinfo{author}{Kanagawa, A.}, \& \bibinfo{author}{Lawi, A.}
  (\bibinfo{year}{2019}).
\newblock \bibinfo{title}{Food constituent estimation for lifestyle disease
  prevention by multi-task {CNN}}.
\newblock {\it \bibinfo{journal}{Applied Artificial Intelligence}\/},  {\it
  \bibinfo{volume}{33}\/}, \bibinfo{pages}{732--746}
  \DOIprefix\doi{https://doi.org/10.1080/08839514.2019.1602318}.
\bibitem[{Slimani et~al.(2000)Slimani, Ferrari, Ocke, Welch, Boeing, Van~Liere,
  Pala, Amiano, Lagiou, Mattisson et~al.}]{slimani2000standardization}
\bibinfo{author}{Slimani, N.}, \bibinfo{author}{Ferrari, P.},
  \bibinfo{author}{Ocke, M.}, \bibinfo{author}{Welch, A.},
  \bibinfo{author}{Boeing, H.}, \bibinfo{author}{Van~Liere, M.},
  \bibinfo{author}{Pala, V.}, \bibinfo{author}{Amiano, P.},
  \bibinfo{author}{Lagiou, A.}, \bibinfo{author}{Mattisson, I.} et~al.
  (\bibinfo{year}{2000}).
\newblock \bibinfo{title}{Standardization of the 24-hour diet recall
  calibration method used in the european prospective investigation into cancer
  and nutrition {(EPIC)}: general concepts and preliminary results}.
\newblock {\it \bibinfo{journal}{European Journal of Clinical Nutrition}\/},
  {\it \bibinfo{volume}{54}\/}, \bibinfo{pages}{900--917}
  \DOIprefix\doi{https://doi.org/10.1038/sj.ejcn.1601107}.
\bibitem[{Snell et~al.(2017)Snell, Swersky \& Zemel}]{DBLP:conf/nips/SnellSZ17}
\bibinfo{author}{Snell, J.}, \bibinfo{author}{Swersky, K.}, \&
  \bibinfo{author}{Zemel, R.~S.} (\bibinfo{year}{2017}).
\newblock \bibinfo{title}{Prototypical networks for few-shot learning}.
\newblock In {\it \bibinfo{booktitle}{Advances in Neural Information Processing
  Systems 30: Annual Conference on Neural Information Processing Systems}\/}
  (pp. \bibinfo{pages}{4077--4087}).
\newblock \URLprefix
  \url{https://proceedings.neurips.cc/paper/2017/hash/cb8da6767461f2812ae4290eac7cbc42-Abstract.html}.
\bibitem[{Subhi et~al.(2019)Subhi, Ali \&
  Mohammed}]{DBLP:journals/access/SubhiAM19}
\bibinfo{author}{Subhi, M.~A.}, \bibinfo{author}{Ali, S.~H.}, \&
  \bibinfo{author}{Mohammed, M.~A.} (\bibinfo{year}{2019}).
\newblock \bibinfo{title}{Vision-based approaches for automatic food
  recognition and dietary assessment: {A} survey}.
\newblock {\it \bibinfo{journal}{{IEEE} Access}\/},  {\it
  \bibinfo{volume}{7}\/}, \bibinfo{pages}{35370--35381}
  \DOIprefix\doi{https://doi.org/10.1109/ACCESS.2019.2904519}.
\bibitem[{Sudo et~al.(2014)Sudo, Murasaki, Shimamura \&
  Taniguchi}]{DBLP:conf/huc/SudoSMT14}
\bibinfo{author}{Sudo, K.}, \bibinfo{author}{Murasaki, K.},
  \bibinfo{author}{Shimamura, J.}, \& \bibinfo{author}{Taniguchi, Y.}
  (\bibinfo{year}{2014}).
\newblock \bibinfo{title}{Estimating nutritional value from food images based
  on semantic segmentation}.
\newblock In {\it \bibinfo{booktitle}{Proceedings of the 2014 ACM International
  Joint Conference on Pervasive and Ubiquitous Computing: Adjunct
  Publication}\/} (pp. \bibinfo{pages}{571--576}).
\newblock \DOIprefix\doi{https://doi.org/10.1145/2638728.2641336}.
\bibitem[{Tahir \& Loo(2021)}]{DBLP:journals/corr/abs-2106-11776}
\bibinfo{author}{Tahir, G.}, \& \bibinfo{author}{Loo, C.~K.}
  (\bibinfo{year}{2021}).
\newblock \bibinfo{title}{A review of the vision-based approaches for dietary
  assessment}.
\newblock {\it \bibinfo{journal}{arXiv preprint}\/},  \URLprefix
  \url{https://arxiv.org/abs/2106.11776}.
\bibitem[{Tay et~al.(2020)Tay, Kaur, Quek, Lim \& Henry}]{tay2020current}
\bibinfo{author}{Tay, W.}, \bibinfo{author}{Kaur, B.}, \bibinfo{author}{Quek,
  R.}, \bibinfo{author}{Lim, J.}, \& \bibinfo{author}{Henry, C.~J.}
  (\bibinfo{year}{2020}).
\newblock \bibinfo{title}{Current developments in digital quantitative volume
  estimation for the optimisation of dietary assessment}.
\newblock {\it \bibinfo{journal}{Nutrients}\/},  {\it \bibinfo{volume}{12}\/},
  \bibinfo{pages}{1167} \DOIprefix\doi{https://doi.org/10.3390/nu12041167}.
\bibitem[{Thames et~al.(2021)Thames, Karpur, Norris, Xia, Panait, Weyand \&
  Sim}]{DBLP:journals/corr/abs-2103-03375}
\bibinfo{author}{Thames, Q.}, \bibinfo{author}{Karpur, A.},
  \bibinfo{author}{Norris, W.}, \bibinfo{author}{Xia, F.},
  \bibinfo{author}{Panait, L.}, \bibinfo{author}{Weyand, T.}, \&
  \bibinfo{author}{Sim, J.} (\bibinfo{year}{2021}).
\newblock \bibinfo{title}{Nutrition5k: Towards automatic nutritional
  understanding of generic food}.
\newblock In {\it \bibinfo{booktitle}{Proceedings of the IEEE/CVF Conference on
  Computer Vision and Pattern Recognition}\/} (pp.
  \bibinfo{pages}{8903--8911}).
\newblock \URLprefix
  \url{https://openaccess.thecvf.com/content/CVPR2021/html/Thames_Nutrition5k_Towards_Automatic_Nutritional_Understanding_of_Generic_Food_CVPR_2021_paper.html}.
\bibitem[{Thompson et~al.(2015)Thompson, Dixit-Joshi, Potischman, Dodd,
  Kirkpatrick, Kushi, Alexander, Coleman, Zimmerman, Sundaram
  et~al.}]{thompson2015comparison}
\bibinfo{author}{Thompson, F.~E.}, \bibinfo{author}{Dixit-Joshi, S.},
  \bibinfo{author}{Potischman, N.}, \bibinfo{author}{Dodd, K.~W.},
  \bibinfo{author}{Kirkpatrick, S.~I.}, \bibinfo{author}{Kushi, L.~H.},
  \bibinfo{author}{Alexander, G.~L.}, \bibinfo{author}{Coleman, L.~A.},
  \bibinfo{author}{Zimmerman, T.~P.}, \bibinfo{author}{Sundaram, M.~E.} et~al.
  (\bibinfo{year}{2015}).
\newblock \bibinfo{title}{Comparison of interviewer-administered and automated
  self-administered 24-hour dietary recalls in 3 diverse integrated health
  systems}.
\newblock {\it \bibinfo{journal}{American Journal of Epidemiology}\/},  {\it
  \bibinfo{volume}{181}\/}, \bibinfo{pages}{970--978}
  \DOIprefix\doi{https://doi.org/10.1093/aje/kwu467}.
\bibitem[{{U.S. Department of Agriculture, Agricultural Research
  Service.}(2020)}]{FNDDS}
\bibinfo{author}{{U.S. Department of Agriculture, Agricultural Research
  Service.}} (\bibinfo{year}{2020}).
\newblock \bibinfo{title}{{USDA} food and nutrient database for dietary studies
  2017-2018}.
\newblock \URLprefix \url{http://www.ars.usda.gov/nea/bhnrc/fsrg}
  \bibinfo{note}{. {Accessed} August 3, 2021}.
\bibitem[{Willett et~al.(1985)Willett, Sampson, Stampfer, Rosner, Bain,
  Witschi, Hennekens \& Speizer}]{willett1985reproducibility}
\bibinfo{author}{Willett, W.~C.}, \bibinfo{author}{Sampson, L.},
  \bibinfo{author}{Stampfer, M.~J.}, \bibinfo{author}{Rosner, B.},
  \bibinfo{author}{Bain, C.}, \bibinfo{author}{Witschi, J.},
  \bibinfo{author}{Hennekens, C.~H.}, \& \bibinfo{author}{Speizer, F.~E.}
  (\bibinfo{year}{1985}).
\newblock \bibinfo{title}{Reproducibility and validity of a semiquantitative
  food frequency questionnaire}.
\newblock {\it \bibinfo{journal}{American Journal of Epidemiology}\/},  {\it
  \bibinfo{volume}{122}\/}, \bibinfo{pages}{51--65}
  \DOIprefix\doi{https://doi.org/10.1093/oxfordjournals.aje.a114086}.
\bibitem[{Wong et~al.(2008)Wong, Boushey, Novotny \&
  Gustafson}]{wong2008evaluation}
\bibinfo{author}{Wong, S.~S.}, \bibinfo{author}{Boushey, C.~J.},
  \bibinfo{author}{Novotny, R.}, \& \bibinfo{author}{Gustafson, D.~R.}
  (\bibinfo{year}{2008}).
\newblock \bibinfo{title}{Evaluation of a computerized food frequency
  questionnaire to estimate calcium intake of asian, hispanic, and non-hispanic
  white youth}.
\newblock {\it \bibinfo{journal}{Journal of the American Dietetic
  Association}\/},  {\it \bibinfo{volume}{108}\/}, \bibinfo{pages}{539--543}
  \DOIprefix\doi{https://doi.org/10.1016/j.jada.2007.12.006}.
\bibitem[{Woo et~al.(2010)Woo, Otsmo, Kim, Ebert, Delp \&
  Boushey}]{DBLP:conf/cimaging/WooOKEDB10}
\bibinfo{author}{Woo, I.}, \bibinfo{author}{Otsmo, K.}, \bibinfo{author}{Kim,
  S.}, \bibinfo{author}{Ebert, D.~S.}, \bibinfo{author}{Delp, E.~J.}, \&
  \bibinfo{author}{Boushey, C.~J.} (\bibinfo{year}{2010}).
\newblock \bibinfo{title}{Automatic portion estimation and visual refinement in
  mobile dietary assessment}.
\newblock In {\it \bibinfo{booktitle}{Computational Imaging VIII}\/} (p.
  \bibinfo{pages}{75330O}).
\newblock \DOIprefix\doi{https://doi.org/10.1117/12.849051}.
\bibitem[{{World Health Organization}(2021)}]{WHO}
\bibinfo{author}{{World Health Organization}} (\bibinfo{year}{2021}).
\newblock \bibinfo{title}{Obesity and overweight}.
\newblock \URLprefix
  \url{https://www.who.int/news-room/fact-sheets/detail/obesity-and-overweight}
  \bibinfo{note}{. {Accessed} December 9, 2021}.
\bibitem[{Wu \& Yang(2009)}]{DBLP:conf/icmcs/WuY09}
\bibinfo{author}{Wu, W.}, \& \bibinfo{author}{Yang, J.} (\bibinfo{year}{2009}).
\newblock \bibinfo{title}{Fast food recognition from videos of eating for
  calorie estimation}.
\newblock In {\it \bibinfo{booktitle}{IEEE International Conference on
  Multimedia and Expo}\/} (pp. \bibinfo{pages}{1210--1213}).
\newblock \DOIprefix\doi{https://doi.org/10.1109/ICME.2009.5202718}.
\bibitem[{Xu et~al.(2013)Xu, He, Khanna, Parra, Boushey \&
  Delp}]{DBLP:conf/mm/XuHKPBD13}
\bibinfo{author}{Xu, C.}, \bibinfo{author}{He, Y.}, \bibinfo{author}{Khanna,
  N.}, \bibinfo{author}{Parra, A.}, \bibinfo{author}{Boushey, C.~J.}, \&
  \bibinfo{author}{Delp, E.~J.} (\bibinfo{year}{2013}).
\newblock \bibinfo{title}{Image-based food volume estimation}.
\newblock In {\it \bibinfo{booktitle}{Proceedings of the 5th International
  Workshop on Multimedia for Cooking {\&} Eating Activities}\/} (pp.
  \bibinfo{pages}{75--80}).
\newblock \DOIprefix\doi{https://doi.org/10.1145/2506023.2506037}.
\bibitem[{Yang et~al.(2019)Yang, Jia, Bucher, Zhang \& Sun}]{yang2019image}
\bibinfo{author}{Yang, Y.}, \bibinfo{author}{Jia, W.}, \bibinfo{author}{Bucher,
  T.}, \bibinfo{author}{Zhang, H.}, \& \bibinfo{author}{Sun, M.}
  (\bibinfo{year}{2019}).
\newblock \bibinfo{title}{Image-based food portion size estimation using a
  smartphone without a fiducial marker}.
\newblock {\it \bibinfo{journal}{Public Health Nutrition}\/},  {\it
  \bibinfo{volume}{22}\/}, \bibinfo{pages}{1180--1192}
  \DOIprefix\doi{https://doi.org/10.1017/S136898001800054X}.
\bibitem[{Yu \& Peng(2006)}]{yu2006robust}
\bibinfo{author}{Yu, C.}, \& \bibinfo{author}{Peng, Q.} (\bibinfo{year}{2006}).
\newblock \bibinfo{title}{Robust recognition of checkerboard pattern for camera
  calibration}.
\newblock {\it \bibinfo{journal}{Optical Engineering}\/},  {\it
  \bibinfo{volume}{45}\/}, \bibinfo{pages}{093201}
  \DOIprefix\doi{https://doi.org/10.1117/1.2352738}.
\bibitem[{Zhang et~al.(2007)Zhang, Wong \&
  Zhang}]{DBLP:journals/pami/ZhangWZ07}
\bibinfo{author}{Zhang, H.}, \bibinfo{author}{Wong, K.~K.}, \&
  \bibinfo{author}{Zhang, G.} (\bibinfo{year}{2007}).
\newblock \bibinfo{title}{Camera calibration from images of spheres}.
\newblock {\it \bibinfo{journal}{IEEE Transactions on Pattern Analysis and
  Machine Intelligence}\/},  {\it \bibinfo{volume}{29}\/},
  \bibinfo{pages}{499--502}
  \DOIprefix\doi{https://doi.org/10.1109/TPAMI.2007.45}.
\bibitem[{Zhou et~al.(2019)Zhou, Zhang, Liu, Qiu \& He}]{zhou2019application}
\bibinfo{author}{Zhou, L.}, \bibinfo{author}{Zhang, C.}, \bibinfo{author}{Liu,
  F.}, \bibinfo{author}{Qiu, Z.}, \& \bibinfo{author}{He, Y.}
  (\bibinfo{year}{2019}).
\newblock \bibinfo{title}{Application of deep learning in food: a review}.
\newblock {\it \bibinfo{journal}{Comprehensive Reviews in Food Science and Food
  Safety}\/},  {\it \bibinfo{volume}{18}\/}, \bibinfo{pages}{1793--1811}
  \DOIprefix\doi{https://doi.org/10.1111/1541-4337.12492}.
\bibitem[{Zhu et~al.(2010{\natexlab{a}})Zhu, Bosch, Boushey \&
  Delp}]{zhu2010image}
\bibinfo{author}{Zhu, F.}, \bibinfo{author}{Bosch, M.},
  \bibinfo{author}{Boushey, C.~J.}, \& \bibinfo{author}{Delp, E.~J.}
  (\bibinfo{year}{2010}{\natexlab{a}}).
\newblock \bibinfo{title}{An image analysis system for dietary assessment and
  evaluation}.
\newblock In {\it \bibinfo{booktitle}{IEEE International Conference on Image
  Processing}\/} (pp. \bibinfo{pages}{1853--1856}).
\newblock \DOIprefix\doi{https://doi.org/10.1109/ICIP.2010.5650848}.
\bibitem[{Zhu et~al.(2015)Zhu, Bosch, Khanna, Boushey \&
  Delp}]{DBLP:journals/titb/ZhuBKBD15}
\bibinfo{author}{Zhu, F.}, \bibinfo{author}{Bosch, M.},
  \bibinfo{author}{Khanna, N.}, \bibinfo{author}{Boushey, C.~J.}, \&
  \bibinfo{author}{Delp, E.~J.} (\bibinfo{year}{2015}).
\newblock \bibinfo{title}{Multiple hypotheses image segmentation and
  classification with application to dietary assessment}.
\newblock {\it \bibinfo{journal}{IEEE Journal of Biomedical and Health
  Informatics}\/},  {\it \bibinfo{volume}{19}\/}, \bibinfo{pages}{377--388}
  \DOIprefix\doi{https://doi.org/10.1109/JBHI.2014.2304925}.
\bibitem[{Zhu et~al.(2010{\natexlab{b}})Zhu, Bosch, Woo, Kim, Boushey, Ebert \&
  Delp}]{DBLP:journals/jstsp/ZhuBWKBED10}
\bibinfo{author}{Zhu, F.}, \bibinfo{author}{Bosch, M.}, \bibinfo{author}{Woo,
  I.}, \bibinfo{author}{Kim, S.}, \bibinfo{author}{Boushey, C.~J.},
  \bibinfo{author}{Ebert, D.~S.}, \& \bibinfo{author}{Delp, E.~J.}
  (\bibinfo{year}{2010}{\natexlab{b}}).
\newblock \bibinfo{title}{The use of mobile devices in aiding dietary
  assessment and evaluation}.
\newblock {\it \bibinfo{journal}{IEEE Journal of Selected Topics in Signal
  Processing}\/},  {\it \bibinfo{volume}{4}\/}, \bibinfo{pages}{756--766}
  \DOIprefix\doi{https://doi.org/10.1109/JSTSP.2010.2051471}.
\bibitem[{Zhu et~al.(2008)Zhu, Mariappan, Boushey, Kerr, Lutes, Ebert \&
  Delp}]{DBLP:conf/cimaging/ZhuMBKLED08}
\bibinfo{author}{Zhu, F.}, \bibinfo{author}{Mariappan, A.},
  \bibinfo{author}{Boushey, C.~J.}, \bibinfo{author}{Kerr, D.},
  \bibinfo{author}{Lutes, K.~D.}, \bibinfo{author}{Ebert, D.~S.}, \&
  \bibinfo{author}{Delp, E.~J.} (\bibinfo{year}{2008}).
\newblock \bibinfo{title}{Technology-assisted dietary assessment}.
\newblock In {\it \bibinfo{booktitle}{Computational Imaging VI}\/} (p.
  \bibinfo{pages}{681411}).
\newblock \DOIprefix\doi{https://doi.org/10.1117/12.778616}.

\end{thebibliography}


\clearpage
\section*{Figure Captions}

\noindent\textbf{Figure \ref{multistage} Multi-stage architecture for dietary assessment.}

\noindent{Multi-stage VBDA architecture generally consists of three stages: food image analysis, portion estimation, and nutrient derivation. Food image analysis is related to food recognition, detection and segmentation. Food portions can be obtained from volume estimates or based on methods such as area. Finally, nutritional information can be obtained from nutrition fact databases.}

\noindent\textbf{Figure \ref{endtoend} End-to-end architecture for dietary assessment.}

\noindent{The end-to-end architecture emphasizes a single model instead of a pipeline of multi-stage methods. Only the original inputs and the final outputs are required to be specified, while the information learned by the neural network is internally relevant.}

\noindent\textbf{Figure \ref{timeline} Representative works for VBDA.}

\noindent{Throughout the history of VBDA, we have summarized  representative studies.}

\noindent\textbf{Figure \ref{work:multi-stage} The framework of multi-stage dietary assessment \citep{DBLP:journals/corr/abs-2003-08273}.}

\noindent{The RGB-D images taken before and after the meal are sequentially analyzed via hyper food semantic segmentation and fine-grained food recognition, volume estimation, and nutrient derivation.}

\noindent\textbf{Figure \ref{work:end-to-end} The framework of MTL end-to-end dietary assessment \citep{DBLP:journals/corr/abs-2103-03375}.}

\noindent{The calorie, macronutrient and mass of the input image are directly estimated  via the designed network with Inception V2 as the main backbone and two 4096-dimensional fully connected (FC) layers. For each task, a final third and fourth FC layers follows and the appropriate loss for the given task is used.}
 \end{document}